\PassOptionsToPackage{table}{xcolor}
\documentclass[10pt]{article} 

\usepackage[preprint]{tmlr}

\usepackage{amsmath,amsfonts,bm}

\def\eqref#1{equation~\ref{#1}}

\def\1{\bm{1}}

\DeclareMathAlphabet{\mathsfit}{\encodingdefault}{\sfdefault}{m}{sl}
\SetMathAlphabet{\mathsfit}{bold}{\encodingdefault}{\sfdefault}{bx}{n}

\usepackage[utf8]{inputenc}
\usepackage{microtype}
\usepackage{xcolor}
\usepackage{hyperref}
\usepackage{url}
\usepackage{booktabs}
\usepackage{graphicx}
\usepackage{multirow}
\usepackage{multicol}
\usepackage{enumitem}
\usepackage{amsmath, amssymb}
\usepackage{algorithm}
\usepackage{algpseudocode}
\usepackage{xspace}
\usepackage{wrapfig}
\usepackage{pifont}

\definecolor{darkblue}{rgb}{0, 0, 0.5}
\definecolor{UpRed}{HTML}{D32F2F}
\definecolor{DownBlue}{HTML}{456A96}

\newcommand{\up}[1]{\,\textcolor{UpRed}{\tiny$\uparrow$\,#1}}
\newcommand{\down}[1]{\,\textcolor{DownBlue}{\tiny$\downarrow$\,#1}}

\hypersetup{colorlinks=true, citecolor=darkblue, linkcolor=darkblue, urlcolor=darkblue}

\newcommand{\added}[1]{#1}

\makeatletter
\if@accepted
  \newcommand{\codelink}{Our code is publicly available at \url{https://github.com/xyliugo/gui-state-transition-pretraining}.}
\else
  \newcommand{\codelink}{Our code is included in the supplementary material and will be released publicly.}
\fi
\makeatother

\title{Scaling GUI Agents with Visual State Transitions}

\author{\textbf{Xiangyan Liu}$^{1}$ \quad \textbf{Kaixin Li}$^{1}$ \quad \textbf{Haonan Wang}$^{1}$ \quad \textbf{Biao Wu}$^{3}$ \quad \textbf{Meng Fang}$^{4}$ \\
\textbf{Longxu Dou}$^{2}$ \quad \textbf{Chao Du}$^{2}$ \quad \textbf{Michael Qizhe Shieh}$^{1}$\thanks{Correspondence to Michael Qizhe Shieh (\href{mailto:michaelshieh@nus.edu.sg}{michaelshieh@nus.edu.sg}).} \quad \textbf{Tianyu Pang}$^{2}$ \\[3pt]
\normalfont\normalsize $^1$NUS \quad $^2$Sea AI Lab \quad $^3$UTS \quad $^4$University of Liverpool}

\def\month{MM}  
\def\year{YYYY} 
\def\openreview{\url{https://openreview.net/forum?id=XXXX}} 

\begin{document}

\maketitle

\begin{abstract}
We introduce State Transition Pretraining (STP) as a new scaling axis for GUI agents. During the STP stage, we continually pretrain a unified multimodal model on visual state transitions by jointly optimizing forward dynamics (predicting next states from current states and actions) and inverse dynamics (predicting actions from state changes). This optimization equips the model with better action-grounded visual representations and an internal world model of GUI dynamics. When subsequently fine-tuned on trajectories with task instructions, our STP-trained models consistently outperform baselines trained solely via direct trajectory fine-tuning across agent benchmarks in both desktop and mobile GUI scenarios (AgentNetBench, AndroidControl, and GUIOdyssey). Further empirical studies show that joint dynamics optimization yields stable improvements over single-objective training, and downstream performance scales steadily with the volume of transition data. \codelink\looseness=-1
\end{abstract}

\begin{figure}[h]
    \centering
    \vspace{-0.0em}
    \includegraphics[width=\linewidth]{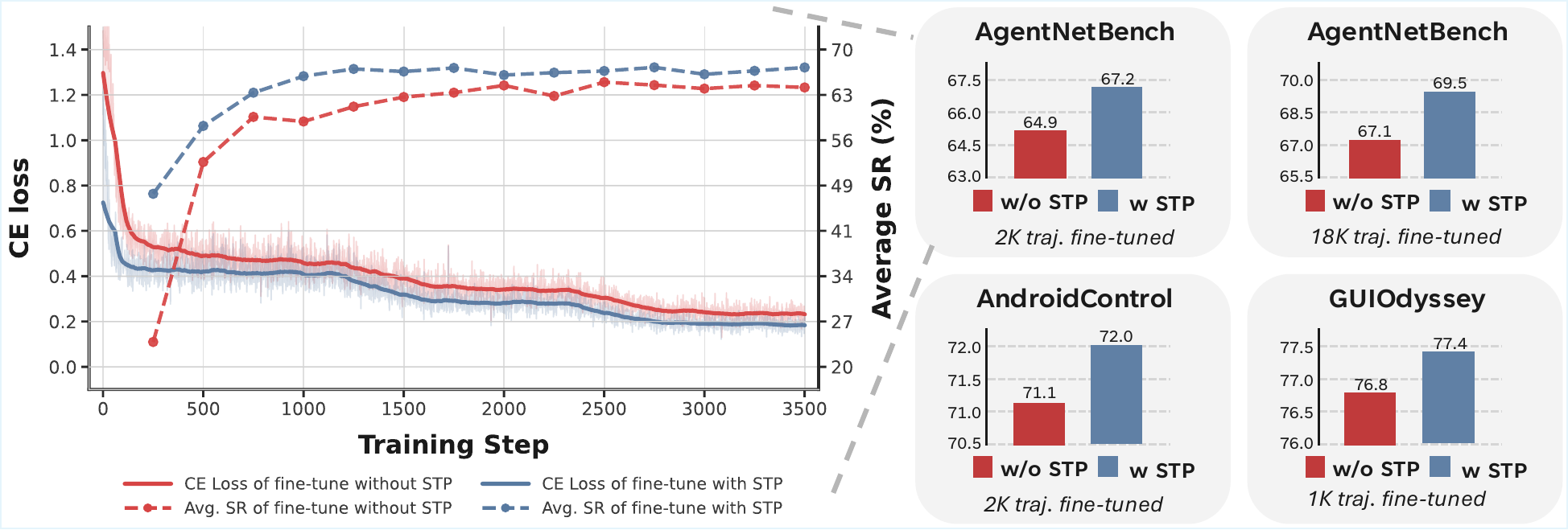}
    \vspace{-0.0em}
    \caption{
        State Transition Pretraining (STP) improves both the training optimization and downstream performance of GUI agents.
        \textbf{Left:} Cross-entropy loss on AgentNet (training) and average success rate on AgentNetBench (evaluation) across fine-tuning steps. Both models are fine-tuned on the exact same 2K AgentNet Win\&Mac trajectories.
        \textbf{Right:} Final performance across multiple benchmarks under different trajectory settings.
        Each subplot corresponds to a distinct training setup, and reports results under identical fine-tuning configurations with and without STP.
        Models are trained and evaluated on the corresponding dataset for each benchmark.
        The y-axis represents average success rate (\%) for AgentNetBench, and step success rate (\%) for AndroidControl and GUIOdyssey.
        These results demonstrate that the STP stage provides a strong initialization that complements trajectory fine-tuning.\looseness=-1
    }
    \vspace{-0.0em}
    \label{fig:teaser}
\end{figure}

\section{Introduction}
\label{sec:intro}
Graphical user interfaces (GUIs) serve as the primary medium for human-software interaction~\citep{myers1998hci-history}. Recently, vision-language models have made it feasible to build GUI agents that automate complex workflows by interpreting task instructions and environment screenshots to predict executable actions~\citep{bai2025qwen3,k25,uitars2,agentnet,scalecua,tang2026clawgui,xu2026mobileagentv35}. Yet, despite their growing practical importance~\citep{operator,cowork,openai2026codexagent}, scaling GUI agents efficiently and effectively remains a significant challenge.

Reinforcement learning (RL) has recently emerged as a common approach for training GUI agents~\citep{ye2025mobile,lai2025computerrl,lu2025arpo,shi2025mobilegui,xu2026mobileagentv35,xue2026evocua}. However, scaling RL poses significant engineering challenges due to the need for complex interactive environments. As a result, trajectory fine-tuning using annotated demonstrations remains a primary approach~\citep{ariaui,aguvis,uitars,yang2025ultracua}, but it still faces limitations that restrict scalability. First, acquiring multi-step trajectory data is expensive because it typically requires manual labeling~\citep{guiodyssey,agentnet,scalecua}. Second, direct fine-tuning on these trajectories entangles visual grounding, dynamics modeling, and task planning into a single objective, making the optimization process difficult.\looseness=-1

We therefore introduce \textbf{State Transition Pretraining (STP)} as a new \emph{scaling axis} for GUI agents to complement trajectory fine-tuning. 
STP requires only step-level visual state transitions, which can be extracted from existing trajectories or collected automatically from interactive environments.
During STP, each transition tuple $(s_t, a_t, s_{t+1})$ is used to instantiate two jointly optimized objectives.
Specifically, \ding{182} \textbf{forward dynamics} operates as an image-text-to-image task that predicts the next state $s_{t+1}$ from the current state and action $(s_t, a_t)$, enabling the model to develop an internal world model of GUI dynamics.
\ding{183} \textbf{inverse dynamics} operates as an \added{image-image-to-text} task that predicts the action $a_t$ from consecutive GUI screenshots $(s_t, s_{t+1})$, encouraging the model to identify \textit{where} the interaction occurred and to extract fine-grained, action-relevant visual features. 
This pretraining stage provides a better initialization for subsequent trajectory fine-tuning.

We evaluate the effectiveness of STP through extensive experiments on desktop and mobile GUI benchmarks. 
To naturally process and generate multimodal inputs and outputs, we adopt a unified multimodal model~\citep{zhao2025unified} as the backbone.
As shown in the main results in Table~\ref{tab:agentnet_main} and Table~\ref{tab:android_main}, when comparing models with and without STP under identical trajectory fine-tuning configurations, we find that STP consistently improves downstream performance across all settings.
Specifically, on AgentNetBench~\citep{agentnet}, STP increases the average success rate by 2.3\% to 6.2\% and the coordinate success rate by 1.6\% to 6.8\%, depending on the fine-tuning data split. In mobile environments, STP yields similar enhancements. On AndroidControl~\citep{androidcontrol}, it improves the step success rate by 0.9\% and the coordinate prediction accuracy by 1.3\%. Similarly, on GUIOdyssey~\citep{guiodyssey}, it increases the step success rate by 0.6\% and the coordinate prediction accuracy by 1.6\%.
In addition, the left panel of Figure~\ref{fig:teaser} illustrates how STP improves the optimization of trajectory fine-tuning as measured by cross-entropy loss.\looseness=-1

Beyond these gains, STP exhibits strong \emph{scalability} properties: scaling the transition pretraining data steadily improves downstream performance when the fine-tuning data is fixed. Moreover, STP continues to outperform fine-tune-only baselines as trajectory fine-tuning data scales from 2K to 18K under a fixed STP setup, suggesting that its gains come not only from additional data exposure but also from the structured learning signals provided by inverse and forward dynamics. Ablation studies further reveal that jointly optimizing inverse and forward dynamics provides complementary supervision, and that STP is almost equally effective whether transitions are sampled by trajectory or as individual steps. A detailed discussion of related work is provided in Section~\ref{sec:related_work}.\looseness=-1
\section{State Transition Pretraining for GUI Agents}
\label{sec:method}
We present \textbf{State Transition Pretraining (STP)}, a pretraining stage for GUI agents built on visual state transitions that require no additional task-level annotations. By modeling local state-action relationships prior to task-level trajectory optimization, it provides a better initialization and serves as a complementary scaling axis to subsequent trajectory fine-tuning.

\begin{figure}[t]
    \centering
    \vspace{-0.0em}
    \includegraphics[width=\linewidth]{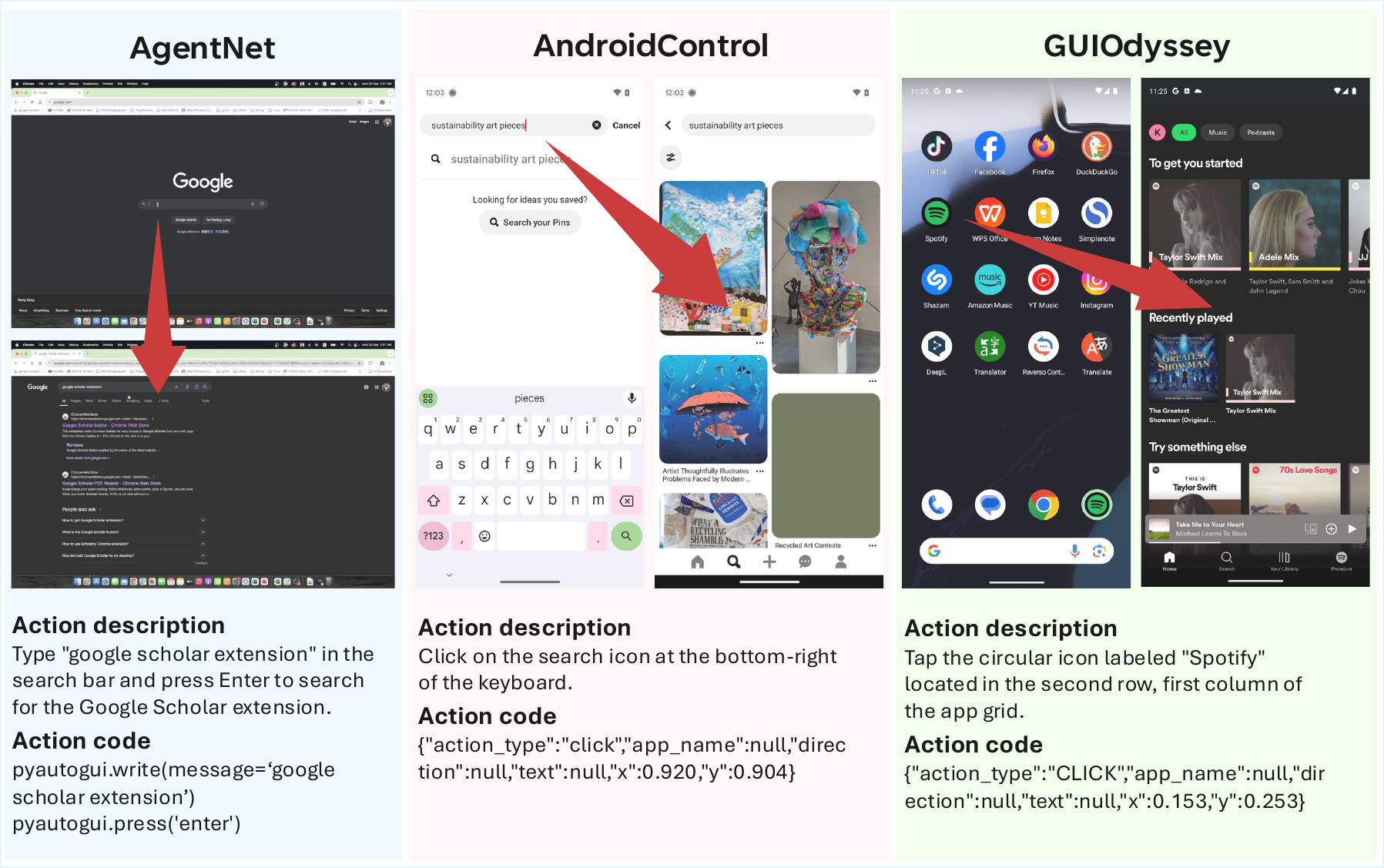}
    \vspace{-0.0em}
    \caption{
    Examples of visual state transition data from three GUI control sources: AgentNet, AndroidControl, and GUIOdyssey.
    Each example consists of a transition tuple $(s_t, a_t, s_{t+1})$, where $s_t$ and $s_{t+1}$ are two consecutive screenshots and $a_t$ is the action executed between them.
    These transitions provide step-level supervision for STP without requiring task-level instruction annotations.
    }
    \vspace{-0.0em}
    \label{fig:transition_sources}
\end{figure}

\subsection{Background}
\label{subsec:background}

\noindent\textbf{Trajectory fine-tuning for GUI agents.}
We formulate GUI control as a sequential decision-making problem based on visual states. Given a task instruction $u$, at step $t$ the agent observes the current GUI screenshot $s_t$ as the state and the interaction history $h_t$, and predicts an action $a_t$ with its specific parameters.
In our setting, $a_t$ is represented as a structured textual output that includes both a natural language description of the interaction and the corresponding executable GUI command (e.g., PyAutoGUI code), rather than solely a discrete action label. 
Here, the interaction history $h_t = \{(s_1, a_1), \dots, (s_{t-1}, a_{t-1})\}$ denotes the sequence of past state–action pairs prior to step $t$. 
This defines a conditional action distribution $p_\theta(a_t \mid u, s_t, h_t)$, 
which is optimized via supervised fine-tuning on expert trajectories:

\begin{equation}
    \mathcal{L}_{\text{SFT}} 
    = - \sum_{t} 
    \log p_\theta(a_t \mid u, s_t, h_t).
    \label{eq:sft}
\end{equation}

However, scaling trajectory fine-tuning with task instructions faces two key challenges:
\begin{itemize}[leftmargin=15pt]
    \item \textbf{Data scarcity:} Collecting high-quality trajectories with accurate task instructions $u$ is expensive and difficult to scale, as existing methods either rely heavily on human annotation~\citep{androidcontrol,guiodyssey,agentnet,scalecua} or use purely synthetic approaches that inevitably introduce noise into the training data~\citep{he2025scalable,feizi2025grounding}.
    \item \textbf{Coupled learning objectives:} Eq.~\ref{eq:sft} implicitly couples multiple capabilities, including visual grounding (identifying and localizing elements in $s_t$), dynamics modeling (understanding action effects), and task planning (following $u$), into a single next-action prediction objective, which can complicate optimization and reduce data efficiency~\citep{seeclick,ni2025rega}.
\end{itemize}

\noindent\textbf{Unified multimodal models.}
Unified multimodal models (UMMs)~\citep{showo,janus,bagel,wu2025harmon} have recently gained increasing attention, as they handle flexible combinations of multimodal inputs and outputs within a single architecture. Specifically, a UMM can accept text, images, or their combination as input and produce text or images as output. This flexibility makes UMMs a natural backbone for state transition pretraining, as the two objectives correspond to native UMM task formats: inverse dynamics maps two screenshots to a textual action description $(s_t, s_{t+1}) \to a_t$, while forward dynamics maps a screenshot and an action to a new screenshot $(s_t, a_t) \to s_{t+1}$, enabling both to be learned jointly within a single model.

\subsection{State Transition Data}
\label{subsec:data}
While trajectory-level annotations are expensive, the step-level state transitions they contain are abundant and annotation-free. We decompose existing trajectory datasets into atomic tuples $(s_t, a_t, s_{t+1})$, each consisting of two consecutive screenshots and the intervening action. Concretely, we extract transitions from AgentNet~\citep{agentnet} for desktop environments, and from AndroidControl~\citep{androidcontrol} and GUIOdyssey~\citep{guiodyssey} for mobile environments, as illustrated in Figure~\ref{fig:transition_sources}.
Because these tuples capture only local state changes and discard task-level instructions, they remain reliable even when the original trajectory labels are noisy, yielding a high-quality transition corpus.\looseness=-1

Given access to interactive environments, programmatic exploration and automated interaction offer a highly scalable method to expand transition collection beyond existing static datasets~\citep{sun2025genesis,ui-oceanus}. Our ablation in Figure~\ref{fig:training_dynamics} (right) shows that individually sampled step-level transitions are as effective as trajectory-level samples under the same transition budget, supporting the viability of this direction. While this presents a highly promising direction, we leave it to future work as it requires substantial engineering resources and falls outside the primary scope of our current study.\looseness=-1

\subsection{Pretraining GUI Agents with Inverse and Forward Dynamics}
\label{subsec:pretraining}
\begin{figure}[t]
    \centering
    \vspace{-0.0em}
    \includegraphics[width=\linewidth]{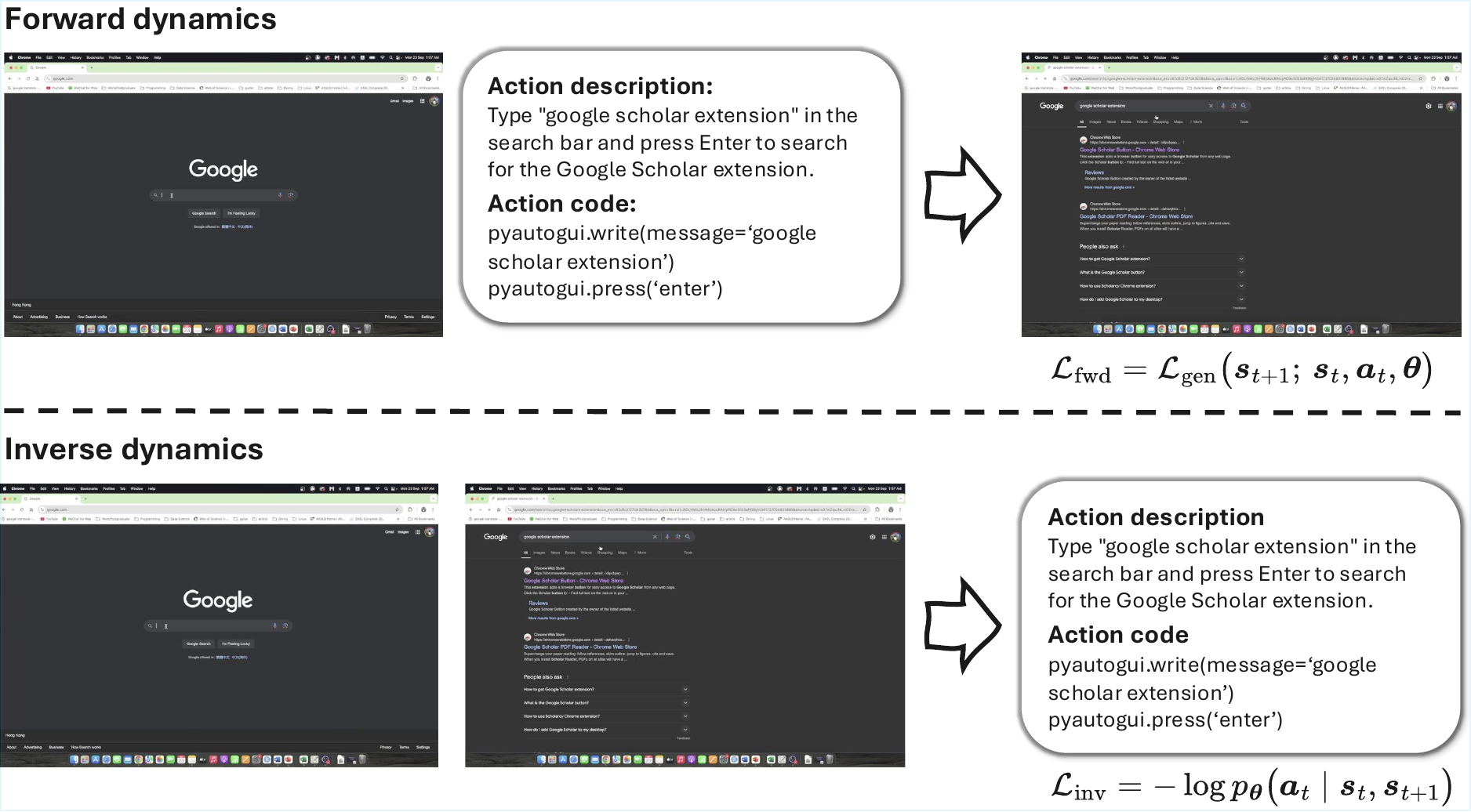}
    \vspace{-0.0em}
    \caption{
    Illustration of inverse and forward dynamics in STP.
    Forward dynamics models the transition $(s_t, a_t) \rightarrow s_{t+1}$ by predicting the next state given the current state and action.
    Inverse dynamics models $(s_t, s_{t+1}) \rightarrow a_t$ by inferring the action that caused the transition.
    }
    \vspace{-0.0em}
    \label{fig:dynamics}
\end{figure}

As shown in Figure~\ref{fig:dynamics}, we continually pretrain the backbone UMM on visual state transitions prior to trajectory fine-tuning using two complementary objectives, inverse dynamics and forward dynamics:
\begin{equation}
    \mathcal{L}_{\text{inv}} = - \log p_\theta(a_t \mid s_t, s_{t+1}), \quad
    \mathcal{L}_{\text{fwd}} = \mathcal{L}_{\text{gen}}(s_{t+1};\, s_t, a_t, \theta),
    \label{eq:dynamics}
\end{equation}
where $p_\theta(a_t \mid s_t, s_{t+1})$ denotes the standard next-token prediction probability for the action given consecutive states, and $\mathcal{L}_{\text{gen}}$ denotes the generative loss optimized by the UMM, which can be instantiated as an autoregressive~\citep{janus}, diffusion-based~\citep{transfusion}, or flow matching~\citep{janusflow,bagel} objective.

\noindent\textbf{Forward dynamics.}
This objective requires the model to predict the next state $s_{t+1}$ given the current state $s_t$ and action $a_t$. Although the optimization is based on a generative loss, recent studies have shown that such generation implicitly provides supervision signals to the model's perception and understanding components, improving their representations~\citep{su2026generation,wu2026visual}. By modeling \textit{what changes} result from an action, the model captures the causal relationship between actions and visual state transitions (e.g., a dropdown list appearing after a click). This encourages the model to internalize how GUIs respond to actions, including changes in element states, content updates, and interface transitions, effectively forming an internal model of GUI dynamics.

\noindent\textbf{Inverse dynamics.}
This objective requires the model to predict the action $a_t$ given the state pair $(s_t, s_{t+1})$. To do so, the model must identify the interaction location and the visual changes caused by the action, such as distinguishing a specific button from similar elements or detecting a text field update, while filtering out irrelevant details. This improves the encoder’s ability to extract fine-grained, action-relevant visual features, while also helping the language modeling component better interpret these representations into accurate action predictions.

\noindent\textbf{Training protocol.} We optimize the joint pretraining objective $\mathcal{L}_{\text{pre}} = \mathcal{L}_{\text{fwd}} + \lambda \mathcal{L}_{\text{inv}}$ on the transition corpus. By deliberately excluding task instructions, this stage enables the model to learn basic GUI capabilities directly from step-level supervision, without the additional complexity of task planning and instruction following. After STP, we perform trajectory fine-tuning to incorporate task-level supervision, allowing the model to integrate these capabilities with instruction-conditioned decision making.\looseness=-1
\section{Experiments}
\label{sec:experiments}

\noindent
In this section, we evaluate State Transition Pretraining (STP) across multiple GUI scenarios, covering different data regimes and training configurations. In particular, we investigate the following aspects:
\begin{itemize}[leftmargin=17pt]
    \item \textbf{Robustness.} STP consistently improves downstream GUI control performance across both desktop and mobile scenarios, covering multiple datasets and benchmarks. (Table~\ref{tab:agentnet_main}, Table~\ref{tab:android_main})

    \item \textbf{Scalability.} Scaling transition data improves performance when fine-tuning data is fixed. Meanwhile, as the amount of fine-tuning data increases, models with STP continue to outperform those without it. (Table~\ref{tab:agentnet_main}, Figure~\ref{fig:ablation_vit_and_pretraining_scale})

    \item \textbf{Impact of pretraining design.} The effectiveness of STP depends on the choice of pretraining objectives and which model components are updated. (Figure~\ref{fig:ablation_inverse_forward_reconstruction}, Figure~\ref{fig:ablation_vit_and_pretraining_scale})
\end{itemize}

\subsection{Experimental Setup}
\label{subsec:experimental_setup}

\textbf{Backbone.} We adopt BAGEL~\citep{bagel}, a unified multimodal model (UMM) that natively supports both text and image generation. BAGEL uses a Mixture-of-Transformers (MoT) architecture containing two specific transformer experts for multimodal understanding and generation. These experts share self-attention layers while maintaining separate parameters for their feed-forward networks. For visual processing, the architecture relies on a Vision Transformer (ViT) as the visual understanding encoder and a Variational Autoencoder (VAE) as the visual generation encoder and decoder. BAGEL uses a flow matching objective for image generation.\looseness=-1

Because of this native structural support for both understanding and generation, BAGEL serves as a natural backbone for our two pretraining objectives: inverse dynamics (image + image $\to$ text) and forward dynamics (image + text $\to$ image). Operating with 7B active parameters per forward pass out of approximately 14B total, BAGEL offers a stronger and more stable foundation to isolate the effect of STP compared with other open-source UMMs such as Show-o~\citep{showo} (1.3B), Harmon~\citep{wu2025harmon} (1.5B), and OpenUni~\citep{openuni} (3.6B).

\textbf{Datasets and evaluation.} We study two representative GUI scenarios: desktop and mobile. For the desktop scenario, we use \textbf{AgentNet}~\citep{agentnet} as the training corpus, which provides 18K Windows and macOS (Win\&Mac) trajectories and 5K Ubuntu trajectories for subsequent trajectory fine-tuning. We further decompose the 18K Win\&Mac trajectories into 320K state transitions for pretraining. We evaluate desktop performance on \textbf{AgentNetBench},\footnote{AgentNetBench is a standalone benchmark built from Win\&Mac samples.} which reports coordinate-, content-, function-level, and average success rates (SRs).
For the mobile scenario, we use datasets from AndroidControl~\citep{androidcontrol} and GUIOdyssey~\citep{guiodyssey}, which provide 12K and 7K trajectories, respectively. From these datasets, we extract 67K and 95K state transitions for pretraining. We measure model performance on AndroidControl~\footnote{We evaluate exclusively on AndroidControl-High because our setting requires generating a low-level instruction before the executable action code (see Figure~\ref{fig:transition_sources}).} and GUIOdyssey using three commonly used metrics for GUI agents: action type prediction accuracy (Type), coordinate prediction accuracy (Grounding), and step success rate (SR).
For all evaluations, we decode the model with temperature 0.6\added{, sample four times per instance, and report the average}.\looseness=-1

\textbf{Comparison setup of main experiments.} Our main comparison is between direct trajectory fine-tuning (\textbf{FT w/o STP}, i.e., fine-tuning without state transition pretraining) and the same fine-tuning pipeline initialized with state transition pretraining (\textbf{FT w/ STP}). For the desktop scenario, we pretrain on 320K AgentNet state transitions and consider three fine-tuning settings: 2K Win\&Mac trajectories, 5K Ubuntu trajectories, and 18K Win\&Mac trajectories. For the mobile scenario, we pretrain on 67K AndroidControl state transitions and fine-tune on 3K AndroidControl trajectories; we also pretrain on 95K GUIOdyssey state transitions and fine-tune on 1K GUIOdyssey trajectories.

We additionally report the performance of several well-known models and checkpoints, including GPT-4o~\citep{gpt4o}, OpenCUA-7B~\citep{agentnet}, and UI-TARS-7B~\citep{uitars}, in the main tables; however, these results are not directly comparable to our method due to differences in training data, backbone architectures, action spaces, and prompting setups across methods, and are therefore included \emph{only for reference}. \added{Our goal is not to claim state-of-the-art performance in absolute terms, but to isolate the effect of the STP stage through controlled comparisons that hold the backbone, fine-tuning data, and training configuration fixed.} Furthermore, our method is largely orthogonal to most existing approaches, meaning it can be applied independently of their specific design choices.

\textbf{Implementation details.} All experiments are conducted on a single node with 8 GPUs, using either NVIDIA A100 40GB GPUs with CPU offloading or NVIDIA H20 96GB GPUs without offloading, depending on hardware availability. We use a learning rate of $1 \times 10^{-5}$ and an EMA decay rate of $0.995$ for both the state transition pretraining (STP) and trajectory fine-tuning stages. During STP, each training step consumes 8 inverse dynamics samples and 8 forward dynamics samples (one of each per GPU); for the joint pretraining objective in Section~\ref{subsec:pretraining}, we set $\lambda=0.25$ for AgentNet and GUIOdyssey, and $\lambda=0.5$ for AndroidControl. During trajectory fine-tuning, we train at step-level granularity following Eq.~\ref{eq:sft}: on AgentNet and GUIOdyssey each GPU processes on average 3 samples per step (24 samples per step in total), and on AndroidControl each GPU processes on average 4 samples per step (32 samples per step in total). We use the same prompt template and input structure for training and inference: the model receives the full action history up to the current step together with only the current observation $s_t$, while past screenshots are excluded from $h_t$ to reduce GPU memory usage. For training on AgentNet, we adopt the L1 (Action) template and do not use the original thought or observation annotations provided in the dataset. For the training of the forward objective, we remove the additional VAE conditioning input in BAGEL. We also note that VAE encoder features are excluded from the input context during training and inference, as they are not used in \added{text-generation} tasks (i.e., inverse dynamics and trajectory fine-tuning). All pretraining on state transitions is conducted for one epoch. \added{For each fine-tuning run, we report the performance of the best-performing checkpoint on the corresponding benchmark, with the identical selection protocol applied to models with and without STP. Concrete examples of how trajectories are converted into fine-tuning and transition samples are provided in Appendix~\ref{sec:sample_construction}.} All other hyperparameters follow the default settings in the BAGEL codebase.\footnote{\url{https://github.com/bytedance-seed/BAGEL}}

\subsection{Main Results}

\begin{table*}[t]
\centering
\scriptsize
\vspace{-0.0em}
\begin{tabular}{lcccc}
\toprule
\textbf{Model}
& \textbf{Coordinate SR (\%)}
& \textbf{Content SR (\%)}
& \textbf{Function SR (\%)}
& \textbf{Average SR (\%)} \\
\midrule
\multicolumn{5}{c}{\textit{Proprietary models}} \\
\midrule
OpenAI CUA~\citep{operator} & 71.7 & 57.3 & 80.0 & 73.1 \\
\midrule
\multicolumn{5}{c}{\textit{Open-source models}} \\
\midrule
Qwen2.5-VL-7B~\citep{qwen25vl} & 50.7 & 40.8 & 3.1 & 48.0 \\
Aguvis-7B~\citep{aguvis} & 56.7 & 43.3 & 0.0 & 52.4 \\
Qwen2.5-VL-32B~\citep{qwen25vl} & 66.6 & 47.2 & 41.5 & 64.8 \\
Qwen2.5-VL-72B~\citep{qwen25vl} & 67.2 & 52.6 & 50.5 & 67.0 \\
OpenCUA-7B~\citep{agentnet} & 79.0 & 62.0 & 44.3 & 75.2 \\
OpenCUA-32B~\citep{agentnet} & 81.9 & 66.1 & 55.7 & 79.1 \\
\midrule
\multicolumn{5}{c}{\textbf{Group 1a}: STP on 320K AgentNet Win\&Mac transitions + FT on 2K AgentNet Win\&Mac trajectories} \\
\midrule
FT w/o STP
& 65.4
& 49.5
& 65.0
& 64.9 \\

\rowcolor{blue!5} FT w/ STP
& 68.0\up{2.6}
& 52.5\up{3.0}
& 69.1\up{4.1}
& 67.2\up{2.3} \\
\midrule
\multicolumn{5}{c}{\textbf{Group 1b}: STP on 320K AgentNet Win\&Mac transitions + FT on 5K AgentNet Ubuntu trajectories} \\
\midrule
FT w/o STP
& 54.2
& 50.2
& 75.3
& 56.9 \\

\rowcolor{blue!5} FT w/ STP
& 61.0\up{6.8}
& 57.6\up{7.4}
& 73.2\down{2.1}
& 63.1\up{6.2} \\
\midrule
\multicolumn{5}{c}{\textbf{Group 1c}: STP on 320K AgentNet Win\&Mac transitions + FT on 18K AgentNet Win\&Mac trajectories} \\
\midrule
FT w/o STP
& 69.4
& 49.3
& 62.9
& 67.1 \\

\rowcolor{blue!5} FT w/ STP
& 71.0\up{1.6}
& 54.4\up{5.1}
& 65.0\up{2.1}
& 69.5\up{2.4} \\
\bottomrule
\end{tabular}
\caption{Main results on AgentNetBench under different fine-tuning regimes. We compare trajectory fine-tuning with and without STP across three settings with varying data scales and domains. STP is performed on 320K AgentNet Win\&Mac state transitions. For FT w/ STP, the colored values indicate the difference relative to FT w/o STP, where red denotes improvement and blue denotes decline.}
\label{tab:agentnet_main}
\vspace{-0.0em}
\end{table*}

\begin{table*}[t]
\centering
\scriptsize
\begin{tabular}{lccccccc}
\toprule
& \multicolumn{3}{c}{\textbf{AndroidControl-High}} & \multicolumn{3}{c}{\textbf{GUIOdyssey}} \\
\cmidrule(lr){2-4} \cmidrule(lr){5-7}
\textbf{Model} & \textbf{Type (\%)} & \textbf{Grounding (\%)} & \textbf{SR (\%)} & \textbf{Type (\%)} & \textbf{Grounding (\%)} & \textbf{SR (\%)}  \\
\midrule
\multicolumn{7}{c}{\textit{Proprietary models}} \\
\midrule
Claude-Computer-Use~\citep{claude-computer-use} & 63.7 & 0.0 & 12.5 & 60.9 & 0.0 & 3.1 \\
GPT-4o~\citep{gpt4o} & 66.3 & 0.0 & 20.8 & 34.3 & 0.0 & 3.3 \\
\midrule
\multicolumn{7}{c}{\textit{Open-source models}} \\
\midrule
SeeClick~\citep{seeclick} & 82.9 & 62.9 & 59.1 & 71.0 & 52.4 & 53.9 \\
OS-Atlas-7B~\citep{wu2024atlas} & 85.2 & 78.5 & 71.2 & 84.5 & 67.8 & 62.0 \\
UI-TARS-7B~\citep{uitars} & 83.7 & 80.5 & 72.5 & 94.6 & 90.1 & 87.0 \\
GUI-R1-7B~\citep{guir1} & 71.6 & 65.6 & 51.7 & 65.5 & 43.6 & 38.8 \\
AgentCPM-GUI~\citep{zhang2025agentcpm} & 77.7 & - & 69.2 & 90.0 & - & 75.0 \\
UI-Venus-Navi-7B~\citep{gu2025ui} & 86.5 & - & 76.1 & 87.3 & - & 71.5 \\
UI-S1-7B~\citep{lu2025ui} & 79.9 & 73.4 & 68.2 & 76.3 & 61.7 & 59.5 \\
GUI-Shift-Mimo-SFT~\citep{uishift} & 87.2 & - & 73.4 & 86.1 & - & 60.7 \\
\midrule
\multicolumn{7}{c}{\textbf{Group 2}: STP on 67K AndroidControl transitions + FT on 3K AndroidControl trajectories} \\
\midrule
FT w/o STP & 86.3 & 68.2 & 71.1 & - & - & - \\
\rowcolor{blue!5} FT w/ STP & 87.1\up{0.8} & 69.5\up{1.3} & 72.0\up{0.9} & - & - & - \\
\midrule
\multicolumn{7}{c}{\textbf{Group 3}: STP on 95K GUIOdyssey transitions + FT on 1K GUIOdyssey trajectories} \\
\midrule
FT w/o STP & - & - & - & 91.3 & 78.8 & 76.8 \\
\rowcolor{blue!5} FT w/ STP & - & - & - & 91.5\up{0.2} & 80.4\up{1.6} & 77.4\up{0.6} \\
\bottomrule
\end{tabular}
\caption{Main results on AndroidControl and GUIOdyssey. For each benchmark, we compare trajectory fine-tuning with and without STP. For FT w/ STP, the colored values indicate the difference relative to FT w/o STP, where red denotes improvement and blue denotes decline.}
\label{tab:android_main}
\vspace{-0.0em}
\end{table*}

\textbf{STP benefits subsequent trajectory fine-tuning across diverse GUI scenarios.}
Tables~\ref{tab:agentnet_main} and~\ref{tab:android_main} demonstrate that, given the same fine-tuning data, models using STP outperform those without it across both desktop and mobile environments. On AgentNetBench (Group 1a), using only 2K Win\&Mac trajectories for fine-tuning, the pretraining stage increases the average SR from 64.9\% to 67.2\% (+2.3\%). Similarly, in mobile scenarios, SR increases from 71.1\% to 72.0\% (+0.9\%) on AndroidControl (Group 2) and from 76.8\% to 77.4\% (+0.6\%) on GUIOdyssey (Group 3). Grounding accuracy also sees improvements of +1.3\% and +1.6\% on AndroidControl and GUIOdyssey, respectively. These gains hold across diverse environments and evaluation metrics, supporting our core claim: STP provides an effective initialization strategy for later trajectory learning across various GUI scenarios.\looseness=-1

\textbf{STP helps under domain-shifted fine-tuning.}
In Group~1b, we pretrain on Win\&Mac transitions but fine-tune on Ubuntu trajectories. This setup introduces a clear domain shift in both the operating system and application distribution, particularly because the target evaluation (AgentNetBench) is based on Win\&Mac environments. Despite this mismatch during the fine-tuning stage, applying STP significantly increases the average SR from 56.9\% to 63.1\% (+6.2\%). This result indicates that target-domain supervision can be effectively supplied through STP rather than relying entirely on trajectory fine-tuning. Consequently, STP offers a practical approach for adapting models to target GUI domains where full trajectory annotation is prohibitively expensive.

\textbf{STP yields improvements even without unseen-data exposure.}
When pretraining improves performance, it is often because the model is exposed to data not seen during the fine-tuning stage. To determine if our results simply come from this additional data exposure, we analyze Group 1c. In this configuration, the 320K state transitions used for STP are extracted directly from the 18K trajectories used for fine-tuning. As a result, the pretraining stage provides no new screenshots, actions, or environment interactions beyond those already used for fine-tuning. Despite this, STP still delivers a performance increase of +2.4\%. This improvement is similar to Group 1a (+2.3\%), where the fine-tuning data is restricted to 2K trajectories and the STP corpus introduces a large amount of additional data. This comparison suggests that the performance gains are not simply driven by exposure to unseen data patterns. Instead, they also arise from the structured learning signals provided by the inverse and forward dynamics objectives.

\begin{figure}[t]
    \centering
    \vspace{-0.0em}
    \begin{minipage}[b]{0.48\linewidth}
        \centering
        \includegraphics[width=\linewidth]{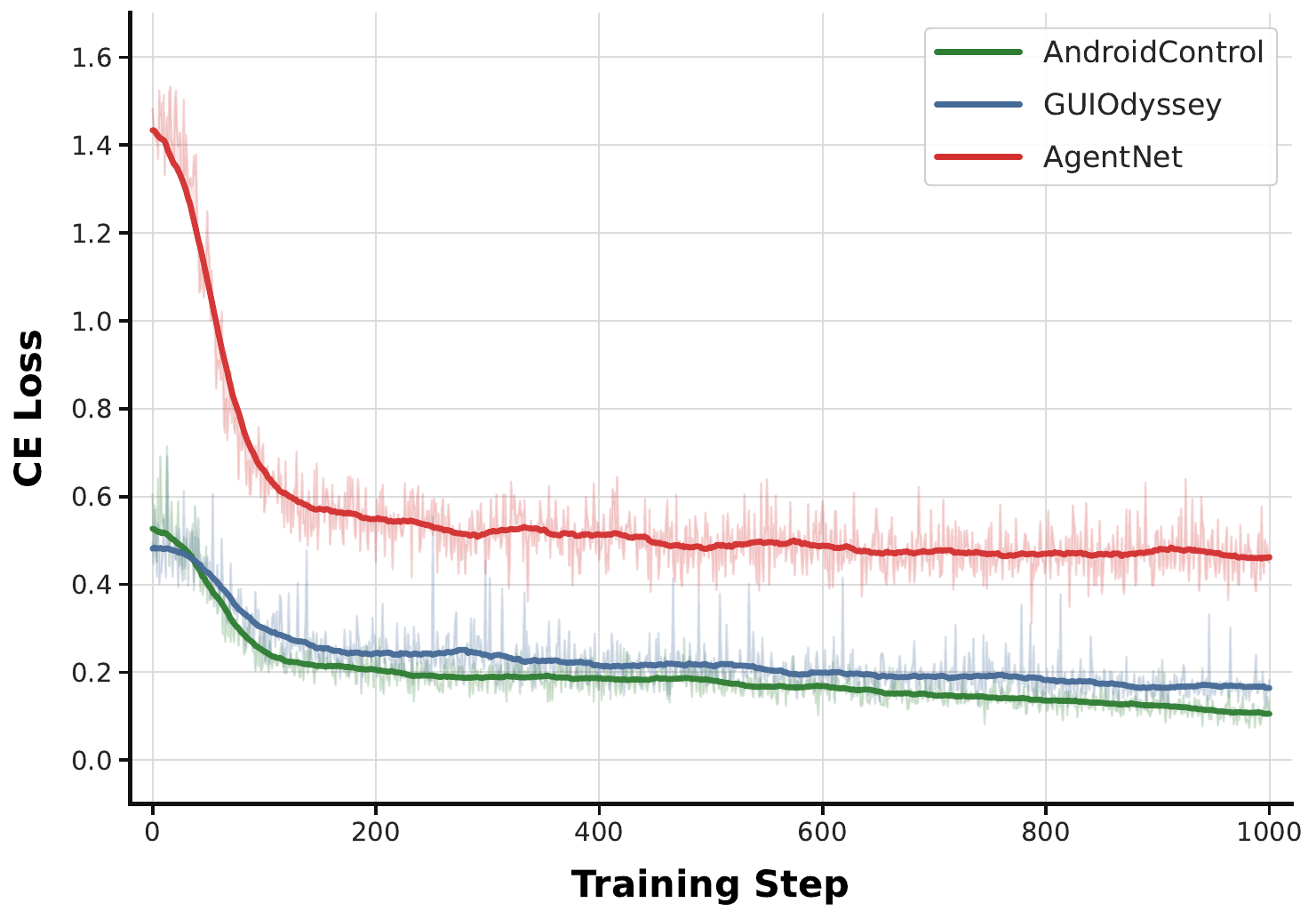}
    \end{minipage}
    \hfill
    \begin{minipage}[b]{0.48\linewidth}
        \centering
        \includegraphics[width=\linewidth]{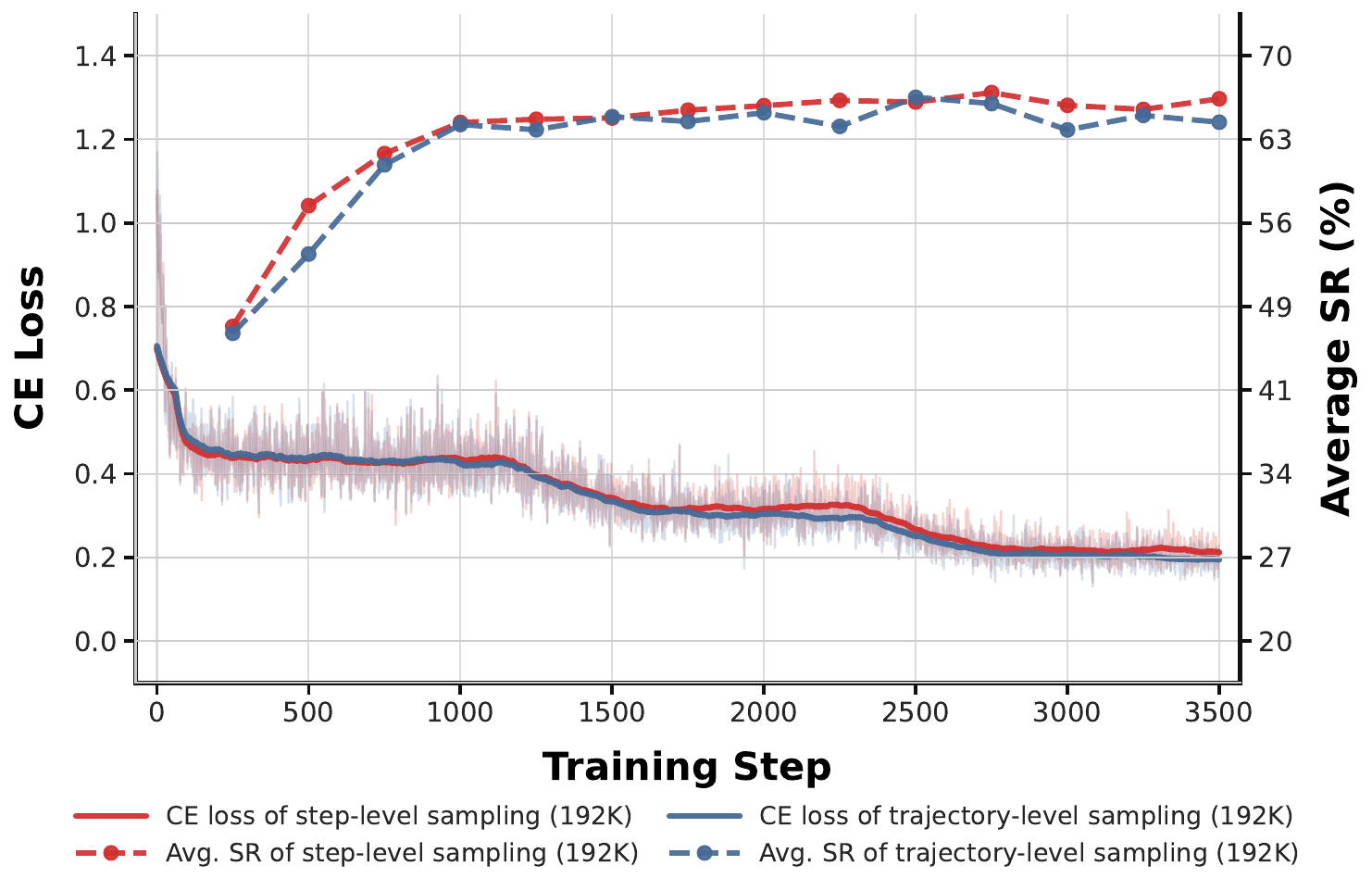}
    \end{minipage}
    \vspace{-0.0em}
    \caption{\added{Training dynamics under different STP conditions. \textbf{Left:} Cross-entropy loss during fine-tuning across the three datasets used in our experiments; mobile datasets (AndroidControl, GUIOdyssey) converge substantially faster than the desktop dataset (AgentNet), indicating lower task complexity and less headroom for STP to fill. \textbf{Right:} Trajectory-level vs.\ step-level transition sampling for STP under a shared 192K-transition budget (fine-tuning on 2K trajectories); the two strategies yield comparable training dynamics and downstream performance.}}
    \label{fig:training_dynamics}
    \vspace{-0.0em}
\end{figure}

\textbf{Smaller gains in mobile scenarios may stem from prior data exposure and lower task complexity.}
While STP improves performance across all evaluated settings, the gains on mobile benchmarks are relatively smaller than those on desktop. We attribute this to two primary factors. First, AndroidControl and GUIOdyssey were released before 2025, so the backbone model (BAGEL), released in May 2025, may have encountered related data during pretraining. This prior exposure reduces the marginal benefit of applying additional STP. Second, mobile tasks in these datasets are generally simpler. This is evident in the training dynamics: as shown in Figure~\ref{fig:training_dynamics} (left), the cross-entropy loss on mobile datasets decreases much faster than on the desktop dataset. Because the model fits the mobile data more easily, there is simply less performance headroom for STP to fill. \added{These observations suggest that the marginal benefit of STP grows with task complexity and with the degree to which the target distribution is under-represented in the backbone's original pretraining data, both of which imply that STP will remain valuable as GUI benchmarks continue to grow in difficulty and diversity.}

\subsection{Ablation Studies}
\label{subsec:ablation}

\begin{figure}[t]
    \centering
    \vspace{-0.0em}
    \includegraphics[width=1\linewidth]{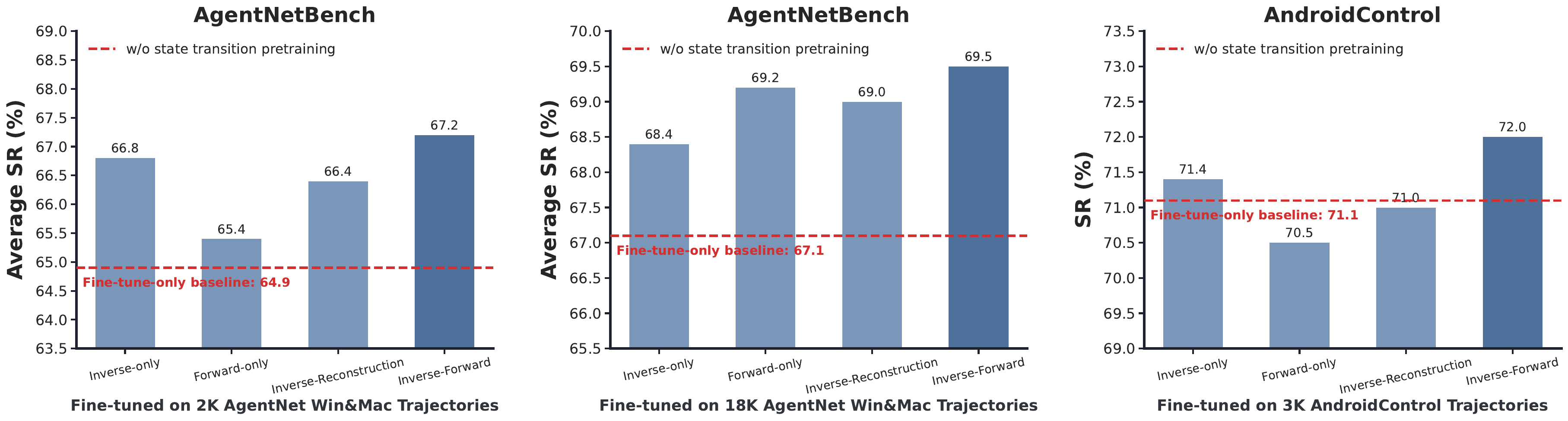}
    \vspace{-0.0em}
    \caption{\textbf{Ablation on pretraining objectives.} We compare \texttt{inverse-only}, \texttt{forward-only}, \texttt{inverse+forward}, and \texttt{inverse+reconstruction} pretraining across different fine-tuning settings. Results are reported as average success rate on AgentNetBench and step accuracy on AndroidControl. Each bar corresponds to a pretraining objective, and the red dashed line indicates the fine-tune-only baseline without state transition pretraining.}
    \label{fig:ablation_inverse_forward_reconstruction}
    \vspace{-0.0em}
\end{figure}

\begin{figure}[t]
    \centering
    \vspace{-0.0em}
    \includegraphics[width=1\linewidth]{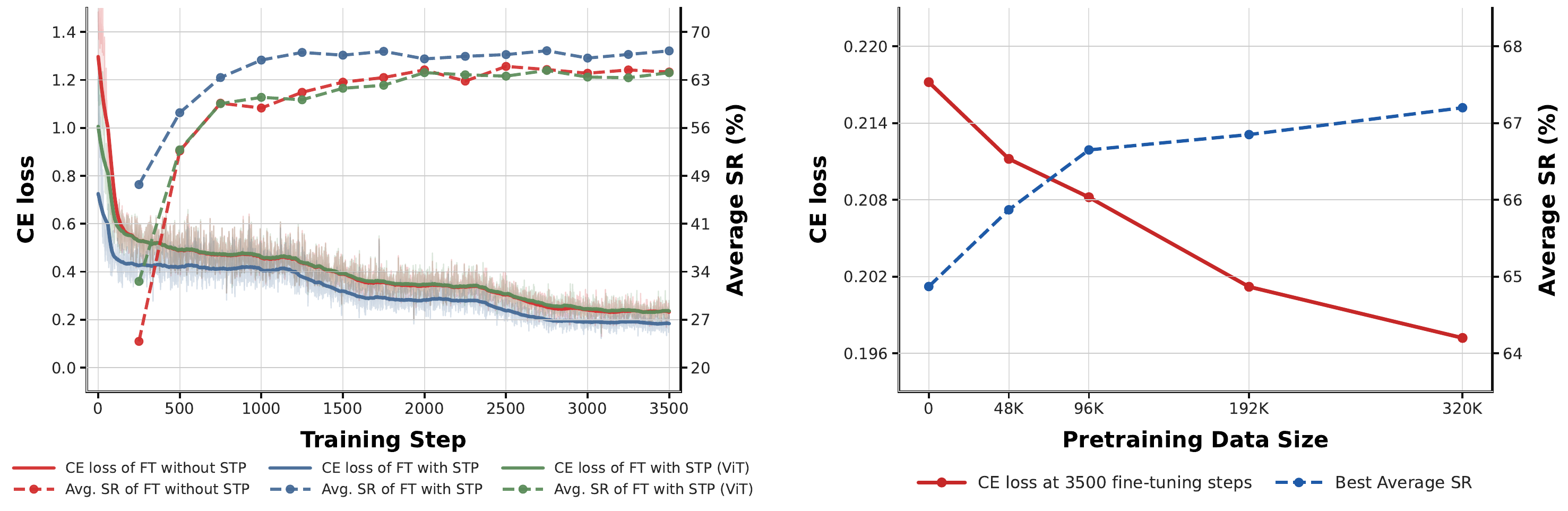}
    \vspace{-0.0em}
    \caption{\textbf{Ablation on visual understanding encoder tuning and pretraining scale.} The left plot compares different pretraining strategies, including no STP, full-model STP, and STP with only the visual understanding encoder (ViT in BAGEL) being updated, with Cross-Entropy (CE) loss and average SR tracked over training steps. The right plot compares different pretraining data sizes, reporting CE loss after 3500 fine-tuning steps and the best average SR achieved under each setting.}
    \label{fig:ablation_vit_and_pretraining_scale}
    \vspace{-0.0em}
\end{figure}

\textbf{Joint inverse and forward dynamics provide the most reliable pretraining signal.}
Figure~\ref{fig:ablation_inverse_forward_reconstruction} compares \texttt{inverse-only}, \texttt{forward-only}, and joint \texttt{inverse+forward} pretraining under different fine-tuning settings. Across all three settings, the joint objective consistently achieves the best performance over the fine-tune-only baseline.
In contrast, the effectiveness of single-objective pretraining is less consistent. On the two desktop settings, both \texttt{inverse-only} and \texttt{forward-only} outperform the fine-tune-only baseline. However, this pattern does not hold on AndroidControl: \texttt{forward-only} drops by -0.6\%, while \texttt{inverse-only} yields only a modest gain of +0.3\%. These results suggest that inverse and forward dynamics provide complementary training signals, and that optimizing them jointly yields a more robust pretraining strategy than relying on either objective alone.\looseness=-1

\textbf{Forward dynamics provides more action-aware supervision than reconstruction.}
As shown in Figure~\ref{fig:ablation_inverse_forward_reconstruction}, \texttt{inverse+forward} consistently outperforms \texttt{inverse+reconstruction} across all settings. The key difference lies in the supervision signal: reconstruction is formulated as generating the same state from $(s_t, \text{``describe this image in detail''}) \rightarrow s_t$, which mainly reinforces static visual understanding and can be solved without modeling action effects. In contrast, forward dynamics learns the transition from $(s_t, a_t)$ to $s_{t+1}$, explicitly capturing how actions change the UI state. When combined with inverse dynamics, this provides more \emph{action-aware} supervision and leads to better alignment with downstream decision making, resulting in consistently stronger performance.

\textbf{Improving visual representations alone is insufficient for GUI control.}
Figure~\ref{fig:ablation_vit_and_pretraining_scale} (left) also compares two pretraining strategies: updating the entire model versus updating only the visual understanding encoder~\footnote{The visual generation encoder and decoder (VAE components) are kept frozen in all settings.}. While tuning only the visual understanding encoder brings a small improvement over no pretraining at the early stage of fine-tuning, the gains quickly diminish, as reflected in both cross-entropy loss and evaluation performance.
In contrast, full-model tuning yields consistent improvements throughout training. This suggests that better visual representations from the ViT alone are not sufficient for strong GUI control. Instead, the language model component plays a critical role in interpreting visual features and producing structured actions, and must be jointly optimized with the visual encoder to fully benefit from STP.

\textbf{Scaling state transition data leads to consistent performance gains.}
As shown in Figure~\ref{fig:ablation_vit_and_pretraining_scale} (right), increasing the amount of transition data used in STP steadily improves downstream performance on AgentNetBench, where the average SR consistently rises as the data scales. Additionally, for a fixed number of fine-tuning steps, models pretrained with more transition data achieve lower cross-entropy loss, indicating a better initialization for downstream optimization.
These findings demonstrate that scaling transition data provides a reliable method for improving GUI agents. This supports our view of STP as an orthogonal scaling axis: larger transition corpora lead to better initialization even when the trajectory fine-tuning budget is fixed. In practice, this approach is highly advantageous because transition data can be obtained at a significantly lower cost than fully annotated trajectories.\looseness=-1

\textbf{Transition sampling granularity has minimal impact on STP effectiveness.}
We compare two transition sampling strategies under the same budget of 192K transitions: (1) \emph{trajectory-level}, where we sample approximately 11K trajectories and decompose them into 192K transitions; and (2) \emph{step-level}, where we directly sample 192K individual transitions from the full 320K AgentNet Win\&Mac transition pool. As shown in Figure~\ref{fig:training_dynamics} (right), the two strategies yield comparable performance (66.85\% vs.\ 66.45\% average SR), despite step-level sampling covering greater data diversity since transitions within the same trajectory tend to share similar applications and visual contexts. This suggests that STP effectiveness is primarily driven by the volume of individual transitions rather than trajectory-level structure, and that future step-level data collected through automated exploration could serve as an effective source.\looseness=-1

\added{\textbf{Practical cost of adding STP.}
A natural concern is whether the gains from STP justify the additional pretraining stage. In our largest configuration (320K AgentNet transitions), STP adds roughly one epoch of joint inverse and forward dynamics training on top of the standard trajectory fine-tuning pipeline, without introducing any additional labels. This cost is small relative to the trajectory-annotation effort that would otherwise be needed to collect a comparable amount of task-level supervision, and it is amortized across all downstream fine-tuning runs that start from the same STP checkpoint. Moreover, all trajectory fine-tuning runs are trained to convergence, as reflected by the plateaued cross-entropy loss in Figure~\ref{fig:training_dynamics} (left), so the gains from STP cannot be recovered by simply allocating more compute to extended fine-tuning. Combined with the observation that transitions can be extracted from the same trajectory pool used for fine-tuning (Group 1c), STP offers a favorable trade-off in practice: a one-time compute investment that yields consistent improvements across multiple downstream configurations without requiring any new human annotation.\looseness=-1}
\section{Related Work}
\label{sec:related_work}

\subsection{GUI Agents}
GUI agents show promise as general interfaces for real-world digital tasks, driving interest in agents that perceive, reason, and act within desktop and mobile environments~\citep{uitars2,zhou2025mai,uivenus1.5,ui-voyager,xiaomigui0}. Existing research covers specific-capability benchmarks~\citep{screenqa,visualwebbench,seeclick,screenspotpro,mu2025gui360,cao2025androidlens} alongside realistic agentic benchmarks featuring longer horizons, richer observations, and complex multi-step interactions~\citep{deng2023mind2web,androidcontrol,androidworld,osworld,guiodyssey,agentnet}. To improve capability, prior work explores supervised fine-tuning (SFT) from expert demonstrations~\citep{aguvis,uitars,agentnet,scalecua} and reinforcement learning (RL) in interactive environments~\citep{ye2025mobile,lai2025computerrl,lu2025arpo,shi2025mobilegui,gu2026generalization,yao2026cgl,mobileforge}. However, scaling these training methods presents significant challenges. RL requires complex environments and extensive engineering support~\citep{mobilegym,phoneworld,weblica}, while SFT relies on multi-step trajectory annotations that are expensive to collect and difficult to scale~\citep{uishift,ui-oceanus}. We introduce State Transition Pretraining (STP) as a new scaling axis for GUI agents. By learning from isolated step-level state transitions that are significantly cheaper and highly scalable, STP provides a strong initialization for subsequent trajectory fine-tuning.

\subsection{Learning from State Transitions}

Recent work has increasingly studied state transitions as a useful learning signal for interactive digital environments. One line of work focuses on \emph{world modeling} for web, mobile, and computer-use agents, where models predict future interface states, state differences, or executable environment representations to support lookahead reasoning and decision making~\citep{chae2024web,vimo,mobiledreamer,zheng2026code2world,koh2026generative,guan2026computer,xu2026mobileworldmodel,patchworld}, interface simulation~\citep{neuralos}, and action refinement \added{or safety checking}~\citep{shen2026world,seerguard}. \added{Most recently, Qwen-AgentWorld~\citep{qwenagentworld} scales this idea into a language world model that simulates diverse agent environments over textual observations, showing that world-model training also serves as an effective warm-up for downstream agent policies.} Another line studies \emph{transition-based supervision} more directly, showing that state changes themselves can provide scalable training signals through self-supervised inverse dynamics~\citep{uishift}, self-supervised forward dynamics or implicit world modeling~\citep{wang2025fostering,cwm,zhang2025agent,chen2025internalizing}, textual foresight~\citep{burns2024tell}, or action recovery from passive videos~\citep{lu2025videoagenttrek,video2gui}. Together, these studies suggest that step-level transitions contain rich information about both action semantics and environment response.

However, existing approaches often treat forward and inverse dynamics separately or rely on indirect supervision. Specifically, UI-Oceanus~\citep{ui-oceanus} demonstrates the scalability of forward dynamics learning but uses textual state differences rather than direct visual prediction as its objective. DeFI~\citep{defi} shows that forward and inverse dynamics provide complementary signals but trains them with separate models rather than jointly. \added{Falcon-UI~\citep{falconui} also decouples instruction-free GUI pretraining from instruction tuning, but focuses on static single-screenshot understanding rather than state transitions.} In contrast, we jointly optimize both objectives within a single unified multimodal model directly over visual states, enabling the two directions to share representations and provide mutual supervision throughout pretraining.

\section{Limitations}
\label{sec:limitations}

Our study has \added{several} limitations. First, most transition data used in our experiments is decomposed from existing trajectory datasets, so the scale and diversity of pretraining remain constrained by the available trajectories. \added{While our step-level vs.\ trajectory-level ablation (Figure~\ref{fig:training_dynamics}, right) suggests that individually collected transitions are equally effective, we have not yet demonstrated this at scales beyond existing trajectory corpora.} Second, although our results indicate that step-level transitions are a promising supervision source, large-scale automated transition collection from interactive environments remains future work and requires substantial engineering effort\added{, including reliable environment setup, action space design, and quality filtering of automatically collected transitions}. Third, the forward dynamics objective introduces additional image-generation pretraining cost \added{compared with text-only inverse dynamics}, which practitioners may need to balance against the downstream gains\added{, particularly under constrained compute budgets}. \added{Fourth, the full STP recipe assumes a backbone with native image-generation capability; on standard vision-language backbones without image generation, only the inverse dynamics objective is directly applicable, a restricted setting that prior work has already explored~\citep{uishift}.}

\section{Broader Impacts}
\label{sec:broader_impacts}

Improving GUI agents may make software more accessible and reduce the cost of automating repetitive digital workflows\added{, which can benefit users with disabilities and lower the barrier to computer use in general}. At the same time, stronger GUI agents could be misused to perform unwanted actions, interact with services without proper authorization, or expose sensitive information if deployed without appropriate safeguards. Responsible deployment should therefore include explicit user consent, sandboxing or permission boundaries, audit logs, and privacy-preserving data handling. \added{Because STP relies only on step-level state transitions and does not require task-level instructions, it avoids one source of sensitive information, as task instructions may embed personal context or private details; however, screenshots themselves can still contain private content and require the same careful data handling.} This paper studies training methods on existing research benchmarks and does not release a new high-risk dataset or an autonomous deployed system.

\section{Conclusion}
\label{sec:conclusion}

We introduce State Transition Pretraining (STP) as a scalable complement to trajectory fine-tuning for GUI agents. By jointly optimizing inverse and forward dynamics on step-level tuples $(s_t, a_t, s_{t+1})$ that can be automatically collected without task annotations, STP consistently improves downstream performance across desktop and mobile benchmarks. Our results further show that the two objectives provide complementary supervision, that performance improves with transition data volume, and that STP remains beneficial as trajectory fine-tuning data scales. \added{Additional analysis reveals that} STP is equally effective whether transitions are decomposed from trajectories or collected at the step level\added{, that improving visual representations alone is insufficient for GUI control, and that the model's language component must be jointly optimized to fully benefit from state-transition supervision. Taken together, these findings position step-level visual state transitions as an underexploited scaling axis for GUI agents, complementary to trajectory fine-tuning and potentially to reinforcement learning, and suggest that future work on large-scale automated transition collection could unlock further gains without incurring the cost of full trajectory annotation}.\looseness=-1

\clearpage
\bibliographystyle{tmlr}
\bibliography{references}

@article{bagel,
  author = {Deng, Chaorui and Zhu, Deyao and Li, Kunchang and Gou, Chenhui and Li, Feng and Wang, Zeyu and Zhong, Shu and Yu, Weihao and Nie, Xiaonan and Song, Ziang and Shi, Guang and Fan, Haoqi},
  title = {Emerging Properties in Unified Multimodal Pretraining},
  journal = {arXiv preprint arXiv:2505.14683},
  year = {2025}
}

@inproceedings{showo,
  title = {Show-o: One Single Transformer to Unify Multimodal Understanding and Generation},
  author = {Xie, Jinheng and Mao, Weijia and Bai, Zechen and Zhang, David Junhao and Wang, Weihao and Lin, Kevin Qinghong and Gu, Yuchao and Chen, Zhijie and Yang, Zhenheng and Shou, Mike Zheng},
  booktitle = {ICLR},
  year = {2025}
}

@article{janus,
  title = {Janus-Pro: Unified Multimodal Understanding and Generation with Data and Model Scaling},
  author = {Chen, Xiaokang and Wu, Zhiyu and Liu, Xingchao and Pan, Zizheng and Liu, Wen and Xie, Zhenda and Yu, Xingkai and Ruan, Chong},
  journal = {arXiv preprint arXiv:2501.17811},
  year = {2025}
}

@article{janusflow,
  title = {{JanusFlow}: Harmonizing Autoregression and Rectified Flow for Unified Multimodal Understanding and Generation},
  author = {Ma, Yiyang and Liu, Xingchao and Chen, Xiaokang and Liu, Wen and Wu, Chengyue and Wu, Zhiyu and Pan, Zizheng and Xie, Zhenda and Zhang, Haowei and Yu, Xingkai and Zhao, Liang and Wang, Yisong and Liu, Jiaying and Ruan, Chong},
  journal = {arXiv preprint arXiv:2411.07975},
  year = {2024}
}

@inproceedings{transfusion,
  title = {Transfusion: Predict the Next Token and Diffuse Images with One Multi-Modal Model},
  author = {Zhou, Chunting and Yu, Lili and Babu, Arun and Tirumala, Kushal and Yasunaga, Michihiro and Shamis, Leonid and Kahn, Jacob and Ma, Xuezhe and Zettlemoyer, Luke and Levy, Omer},
  booktitle = {ICLR},
  year = {2025}
}

@article{openuni,
  title = {OpenUni: A Simple Baseline for Unified Multimodal Understanding and Generation},
  author = {Wu, Size and Wu, Zhonghua and Gong, Zerui and Tao, Qingyi and Jin, Sheng and Li, Qinyue and Li, Wei and Loy, Chen Change},
  journal = {arXiv preprint arXiv:2505.23661},
  year = {2025}
}

@article{wu2025harmon,
  title = {Harmonizing Visual Representations for Unified Multimodal Understanding and Generation},
  author = {Wu, Size and Zhang, Wenwei and Xu, Lumin and Jin, Sheng and Wu, Zhonghua and Tao, Qingyi and Liu, Wentao and Li, Wei and Loy, Chen Change},
  journal = {arXiv preprint arXiv:2503.21979},
  year = {2025}
}

@inproceedings{seeclick,
  author = {Cheng, Kanzhi and Sun, Qiushi and Chu, Yougang and Xu, Fangzhi and Li, Yantao and Zhang, Jianbing and Wu, Zhiyong},
  title = {{SeeClick}: Harnessing {GUI} Grounding for Advanced Visual {GUI} Agents},
  booktitle = {Proceedings of the 62nd Annual Meeting of the Association for Computational Linguistics (ACL)},
  year = {2024},
  note = {arXiv:2401.10935}
}

@article{uitars,
  author = {Qin, Yujia and Ye, Yining and Fang, Junjie and Wang, Haoming and Liang, Shihao and Tian, Shizuo and Zhang, Junda and Li, Jiahao and Li, Yunxin and Huang, Shijue and others},
  title = {{UI-TARS}: Pioneering Automated {GUI} Interaction with Native Agents},
  journal = {arXiv preprint arXiv:2501.12326},
  year = {2025}
}

@article{ariaui,
  author = {Yang, Yuhao and Wang, Yue and Li, Dongxu and Luo, Ziyang and Chen, Bei and Huang, Chao and Li, Junnan},
  title = {{Aria-UI}: Visual Grounding for {GUI} Instructions},
  journal = {arXiv preprint arXiv:2412.16256},
  year = {2024}
}

@article{uitars2,
  author = {Wang, Haoming and Zou, Haoyang and Song, Huatong and Feng, Jiazhan and Fang, Junjie and others},
  title = {{UI-TARS-2} Technical Report: Advancing {GUI} Agent with Multi-Turn Reinforcement Learning},
  journal = {arXiv preprint arXiv:2509.02544},
  year = {2025}
}

@inproceedings{osworld,
  author = {Xie, Tianbao and Zhang, Danyang and Chen, Jixuan and Li, Xiaochuan and Zhao, Siheng and Cao, Ruisheng and Hua, Toh Jing and Cheng, Zhoujun and Shin, Dongchan and Lei, Fangyu and Liu, Yitao and Xu, Yiheng and Zhou, Shuyan and Savarese, Silvio and Xiong, Caiming and Zhong, Victor and Yu, Tao},
  title = {{OSWorld}: Benchmarking Multimodal Agents for Open-Ended Tasks in Real Computer Environments},
  booktitle = {Advances in Neural Information Processing Systems (NeurIPS)},
  year = {2024},
  note = {arXiv:2404.07972}
}

@misc{agentnet,
  author = {Wang, Xinyuan and Wang, Bowen and Lu, Dunjie and Yang, Junlin and Xie, Tianbao and Wang, Junli and Deng, Jiaqi and Guo, Xiaole and Xu, Yiheng and Wu, Chen Henry and Shen, Zhennan and Li, Zhuokai and Li, Ryan and Li, Xiaochuan and Chen, Junda and Zheng, Boyuan and Li, Peihang and Lei, Fangyu and Cao, Ruisheng and Fu, Yeqiao and Shin, Dongchan and Shin, Martin and Hu, Jiarui and Wang, Yuyan and Chen, Jixuan and Ye, Yuxiao and Zhang, Danyang and Du, Dikang and Hu, Hao and Chen, Huarong and Zhou, Zaida and Yao, Haotian and Chen, Ziwei and Gu, Qizheng and Wang, Yipu and Wang, Heng and Yang, Diyi and Zhong, Victor and Sung, Flood and Charles, Y. and Yang, Zhilin and Yu, Tao},
  title = {{OpenCUA}: Open Foundations for Computer-Use Agents},
  howpublished = {\url{https://opencua.xlang.ai/}},
  year = {2025},
  note = {AgentNet dataset and AgentNetBench described on project page},
  eprint = {2508.09123},
  archivePrefix = {arXiv},
  primaryClass = {cs.AI}
}

@inproceedings{androidcontrol,
  author = {Li, Wei and Bishop, William and Li, Alice and Rawles, Chris and Campbell-Ajala, Folawiyo and Tyamagundlu, Divya and Riva, Oriana},
  title = {On the Effects of Data Scale on {UI} Control Agents},
  booktitle = {Advances in Neural Information Processing Systems (NeurIPS)},
  year = {2024},
  note = {arXiv:2406.03679}
}

@article{uishift,
  author = {Gao, Longxi and Zhang, Li and Gao, Pengzhi and Liu, Wei and Luan, Jian and Xu, Mengwei},
  title = {{GUI-Shift}: Enhancing {VLM}-based {GUI} Agents through Self-supervised Reinforcement Learning},
  journal = {arXiv preprint arXiv:2505.12493},
  year = {2025}
}

@article{vimo,
  author = {Luo, Dezhao and Tang, Bohan and Li, Kang and Papoudakis, Georgios and Song, Jifei and Gong, Shaogang and Hao, Jianye and Wang, Jun and Shao, Kun},
  title = {{ViMo}: A Generative Visual {GUI} World Model for App Agents},
  journal = {arXiv preprint arXiv:2504.13936},
  year = {2025}
}

@article{mobiledreamer,
  author = {Cao, Yilin and Zhong, Yufeng and Zeng, Zhixiong and Zheng, Liming and Huang, Jing and Qiu, Haibo and Shi, Peng and Mao, Wenji and Guanglu, Wan},
  title = {{MobileDreamer}: Generative Sketch World Model for {GUI} Agent},
  journal = {arXiv preprint arXiv:2601.04035},
  year = {2026}
}

@article{neuralos,
  author = {Rivard, Luke and Sun, Sun and Guo, Hongyu and Chen, Wenhu and Deng, Yuntian},
  title = {{NeuralOS}: Towards Simulating Operating Systems via Neural Generative Models},
  journal = {arXiv preprint arXiv:2507.08800},
  year = {2025},
  eprint = {2507.08800},
  archivePrefix = {arXiv},
  primaryClass = {cs.AI}
}

@inproceedings{
defi,
title={Disentangled Robot Learning via Separate Forward and Inverse Dynamics Pretraining},
author={Wenyao Zhang and Bozhou Zhang and Zekun Qi and Wenjun Zeng and Xin Jin and Li Zhang},
booktitle={The Fourteenth International Conference on Learning Representations},
year={2026},
url={https://openreview.net/forum?id=DdrsHWobR1}
}

@article{guir1,
  author = {Luo, Run and Wang, Lu and He, Wanwei and Chen, Longze and Li, Jiaming and Xia, Xiaobo},
  title = {{GUI-R1}: A Generalist {R1}-Style Vision-Language Action Model for {GUI} Agents},
  journal = {arXiv preprint arXiv:2504.10458},
  year = {2025}
}

@article{qwen25vl,
  title={{Qwen2.5-VL} technical report},
  author={Bai, Shuai and Chen, Keqin and Liu, Xuejing and Wang, Jialin and Ge, Wenbin and Song, Sibo and Dang, Kai and Wang, Peng and Wang, Shijie and Tang, Jun and others},
  journal={arXiv preprint arXiv:2502.13923},
  year={2025}
}

@article{k25,
  title={{Kimi K2.5}: Visual Agentic Intelligence},
  author={Team, Kimi and Bai, Tongtong and Bai, Yifan and Bao, Yiping and Cai, SH and Cao, Yuan and Charles, Y and Che, HS and Chen, Cheng and Chen, Guanduo and others},
  journal={arXiv preprint arXiv:2602.02276},
  year={2026}
}

@article{scalecua,
  title={{ScaleCUA}: Scaling open-source computer use agents with cross-platform data},
  author={Liu, Zhaoyang and Xie, JingJing and Ding, Zichen and Li, Zehao and Yang, Bowen and Wu, Zhenyu and Wang, Xuehui and Sun, Qiushi and Liu, Shi and Wang, Weiyun and others},
  journal={arXiv preprint arXiv:2509.15221},
  year={2025}
}

@inproceedings{guiodyssey,
  title={{GUIOdyssey}: A comprehensive dataset for cross-app GUI navigation on mobile devices},
  author={Lu, Quanfeng and Shao, Wenqi and Liu, Zitao and Du, Lingxiao and Meng, Fanqing and Li, Boxuan and Chen, Botong and Huang, Siyuan and Zhang, Kaipeng and Luo, Ping},
  booktitle={Proceedings of the IEEE/CVF International Conference on Computer Vision},
  pages={22404--22414},
  year={2025}
}

@article{aguvis,
  title={Aguvis: Unified pure vision agents for autonomous gui interaction},
  author={Xu, Yiheng and Wang, Zekun and Wang, Junli and Lu, Dunjie and Xie, Tianbao and Saha, Amrita and Sahoo, Doyen and Yu, Tao and Xiong, Caiming},
  journal={arXiv preprint arXiv:2412.04454},
  year={2024}
}

@article{cwm,
  title={{CWM}: An open-weights LLM for research on code generation with world models},
  author={{FAIR CodeGen team} and Copet, Jade and Carbonneaux, Quentin and Cohen, Gal and Gehring, Jonas and Kahn, Jacob and Kossen, Jannik and Kreuk, Felix and McMilin, Emily and Meyer, Michel and Wei, Yuxiang and others},
  journal={arXiv preprint arXiv:2510.02387},
  year={2025}
}

@article{zhang2025agent,
  title={Agent learning via early experience},
  author={Zhang, Kai and Chen, Xiangchao and Liu, Bo and Xue, Tianci and Liao, Zeyi and Liu, Zhihan and Wang, Xiyao and Ning, Yuting and Chen, Zhaorun and Fu, Xiaohan and others},
  journal={arXiv preprint arXiv:2510.08558},
  year={2025}
}

@misc{operator,
  author       = {{OpenAI}},
  title        = {Introducing Operator},
  year         = {2025},
  howpublished = {\url{https://openai.com/index/introducing-operator/}},
  note         = {Product page, accessed April 3, 2026}
}

@misc{cowork,
  author       = {{Anthropic}},
  title        = {Claude Cowork},
  year         = {2026},
  howpublished = {\url{https://www.anthropic.com/product/claude-cowork}},
  note         = {Product page, accessed April 3, 2026}
}

@article{feizi2025grounding,
  title={Grounding Computer Use Agents on Human Demonstrations},
  author={Feizi, Aarash and Nayak, Shravan and Jian, Xiangru and Lin, Kevin Qinghong and Li, Kaixin and Awal, Rabiul and L{\`u}, Xing Han and Obando-Ceron, Johan and Rodriguez, Juan A and Chapados, Nicolas and others},
  journal={arXiv preprint arXiv:2511.07332},
  year={2025}
}

@article{he2025scalable,
  title={{WebSTAR}: Scalable Data Synthesis for Computer Use Agents with Step-Level Filtering},
  author={He, Yifei and Chawla, Pranit and Souri, Yaser and Som, Subhojit and Song, Xia},
  journal={arXiv preprint arXiv:2512.10962},
  year={2025}
}

@inproceedings{ni2025rega,
  title={ReGA: Reasoning and Grounding Decoupled GUI Navigation Agents},
  author={Ni, Feiyue and Guan, Yanchu and Sun, Yuchong and Wang, Dong and Zhuang, Chenyi and Gu, Jinjie and Song, Ruihua},
  booktitle={CCF International Conference on Natural Language Processing and Chinese Computing},
  pages={375--387},
  year={2025},
  organization={Springer}
}

@article{su2026generation,
  title={Generation Enhances Understanding in Unified Multimodal Models via Multi-Representation Generation},
  author={Su, Zihan and Wei, Hongyang and Cen, Kangrui and Wang, Yong and Chen, Guanhua and Yuan, Chun and Chu, Xiangxiang},
  journal={arXiv preprint arXiv:2601.21406},
  year={2026}
}

@article{wu2026visual,
  title={Visual Generation Unlocks Human-Like Reasoning through Multimodal World Models},
  author={Wu, Jialong and Zhang, Xiaoying and Yuan, Hongyi and Zhang, Xiangcheng and Huang, Tianhao and He, Changjing and Deng, Chaoyi and Zhang, Renrui and Wu, Youbin and Long, Mingsheng},
  journal={arXiv preprint arXiv:2601.19834},
  year={2026}
}

@article{gpt4o,
  title        = {GPT-4o System Card},
  author       = {{OpenAI}},
  year         = {2024},
  journal      = {arXiv preprint arXiv:2410.21276},
  url          = {https://arxiv.org/abs/2410.21276}
}

@article{ui-voyager,
  title={UI-Voyager: A Self-Evolving GUI Agent Learning via Failed Experience},
  author={Lin, Zichuan and Liu, Feiyu and Yang, Yijun and Lyu, Jiafei and Gao, Yiming and Liu, Yicheng and Lu, Zhicong and Yu, Yangbin and Yang, Mingyu and Li, Junyou and others},
  journal={arXiv preprint arXiv:2603.24533},
  year={2026}
}

@article{uivenus1.5,
  title={UI-Venus-1.5 Technical Report},
  author={Team, Venus and Gao, Changlong and Gu, Zhangxuan and Liu, Yulin and Qiu, Xinyu and Shen, Shuheng and Wen, Yue and Xia, Tianyu and Xu, Zhenyu and Zeng, Zhengwen and others},
  journal={arXiv preprint arXiv:2602.09082},
  year={2026}
}

@inproceedings{screenqa,
  title={Screenqa: Large-scale question-answer pairs over mobile app screenshots},
  author={Hsiao, Yu-Chung and Zubach, Fedir and Baechler, Gilles and Sunkara, Srinivas and C{\u{a}}rbune, Victor and Lin, Jason and Wang, Maria and Zhu, Yun and Chen, Jindong},
  booktitle={Proceedings of the 2025 Conference of the Nations of the Americas Chapter of the Association for Computational Linguistics: Human Language Technologies (Volume 1: Long Papers)},
  pages={9427--9452},
  year={2025}
}

@inproceedings{screenspotpro,
  title={Screenspot-pro: Gui grounding for professional high-resolution computer use},
  author={Li, Kaixin and Meng, Ziyang and Lin, Hongzhan and Luo, Ziyang and Tian, Yuchen and Ma, Jing and Huang, Zhiyong and Chua, Tat-Seng},
  booktitle={Proceedings of the 33rd ACM International Conference on Multimedia},
  pages={8778--8786},
  year={2025}
}

@article{gu2025ui,
  title={Ui-venus technical report: Building high-performance ui agents with rft},
  author={Gu, Zhangxuan and Zeng, Zhengwen and Xu, Zhenyu and Zhou, Xingran and Shen, Shuheng and Liu, Yunfei and Zhou, Beitong and Meng, Changhua and Xia, Tianyu and Chen, Weizhi and others},
  journal={arXiv preprint arXiv:2508.10833},
  year={2025}
}

@article{visualwebbench,
  title={Visualwebbench: How far have multimodal llms evolved in web page understanding and grounding?},
  author={Liu, Junpeng and Song, Yifan and Lin, Bill Yuchen and Lam, Wai and Neubig, Graham and Li, Yuanzhi and Yue, Xiang},
  journal={arXiv preprint arXiv:2404.05955},
  year={2024}
}

@article{deng2023mind2web,
  title={Mind2web: Towards a generalist agent for the web},
  author={Deng, Xiang and Gu, Yu and Zheng, Boyuan and Chen, Shijie and Stevens, Samuel and Wang, Boshi and Sun, Huan and Su, Yu},
  journal={Advances in Neural Information Processing Systems},
  volume={36},
  pages={28091--28114},
  year={2023}
}

@article{androidworld,
  title={Androidworld: A dynamic benchmarking environment for autonomous agents},
  author={Rawles, Christopher and Clinckemaillie, Sarah and Chang, Yifan and Waltz, Jonathan and Lau, Gabrielle and Fair, Marybeth and Li, Alice and Bishop, William and Li, Wei and Campbell-Ajala, Folawiyo and others},
  journal={arXiv preprint arXiv:2405.14573},
  year={2024}
}

@article{ye2025mobile,
  title={Mobile-agent-v3: Fundamental agents for gui automation},
  author={Ye, Jiabo and Zhang, Xi and Xu, Haiyang and Liu, Haowei and Wang, Junyang and Zhu, Zhaoqing and Zheng, Ziwei and Gao, Feiyu and Cao, Junjie and Lu, Zhengxi and others},
  journal={arXiv preprint arXiv:2508.15144},
  year={2025}
}

@article{lu2025arpo,
  title={Arpo: End-to-end policy optimization for gui agents with experience replay},
  author={Lu, Fanbin and Zhong, Zhisheng and Liu, Shu and Fu, Chi-Wing and Jia, Jiaya},
  journal={arXiv preprint arXiv:2505.16282},
  year={2025}
}

@article{ui-oceanus,
  title={UI-Oceanus: Scaling GUI Agents with Synthetic Environmental Dynamics},
  author={Wu, Mengzhou and Guo, Yuzhe and Cao, Yuan and Lu, Haochuan and Zhu, Songhe and Qu, Pingzhe and Chen, Xin and Qin, Kang and Wang, Zhongpu and Zhang, Xiaode and others},
  journal={arXiv preprint arXiv:2604.02345},
  year={2026}
}

@article{zhou2025mai,
  title={MAI-UI Technical Report: Real-World Centric Foundation GUI Agents},
  author={Zhou, Hanzhang and Zhang, Xu and Tong, Panrong and Zhang, Jianan and Chen, Liangyu and Kong, Quyu and Cai, Chenglin and Liu, Chen and Wang, Yue and Zhou, Jingren and others},
  journal={arXiv preprint arXiv:2512.22047},
  year={2025}
}

@article{shi2025mobilegui,
  title={Mobilegui-rl: Advancing mobile gui agent through reinforcement learning in online environment},
  author={Shi, Yucheng and Yu, Wenhao and Li, Zaitang and Wang, Yonglin and Zhang, Hongming and Liu, Ninghao and Mi, Haitao and Yu, Dong},
  journal={arXiv preprint arXiv:2507.05720},
  year={2025}
}

@article{lai2025computerrl,
  title={Computerrl: Scaling end-to-end online reinforcement learning for computer use agents},
  author={Lai, Hanyu and Liu, Xiao and Zhao, Yanxiao and Xu, Han and Zhang, Hanchen and Jing, Bohao and Ren, Yanyu and Yao, Shuntian and Dong, Yuxiao and Tang, Jie},
  journal={arXiv preprint arXiv:2508.14040},
  year={2025}
}

@article{chae2024web,
  title={Web agents with world models: Learning and leveraging environment dynamics in web navigation},
  author={Chae, Hyungjoo and Kim, Namyoung and Ong, Kai Tzu-iunn and Gwak, Minju and Song, Gwanwoo and Kim, Jihoon and Kim, Sunghwan and Lee, Dongha and Yeo, Jinyoung},
  journal={arXiv preprint arXiv:2410.13232},
  year={2024}
}

@inproceedings{burns2024tell,
  title={Tell me what’s next: Textual foresight for generic ui representations},
  author={Burns, Andrea and Saenko, Kate and Plummer, Bryan},
  booktitle={Findings of the Association for Computational Linguistics: ACL 2024},
  pages={4590--4611},
  year={2024}
}

@article{zheng2026code2world,
  title={Code2world: A gui world model via renderable code generation},
  author={Zheng, Yuhao and Zhong, Li'an and Wang, Yi and Dai, Rui and Liu, Kaikui and Chu, Xiangxiang and Lv, Linyuan and Torr, Philip and Lin, Kevin Qinghong},
  journal={arXiv preprint arXiv:2602.09856},
  year={2026}
}

@article{lu2025videoagenttrek,
  title={VideoAgentTrek: Computer Use Pretraining from Unlabeled Videos},
  author={Lu, Dunjie and Xu, Yiheng and Wang, Junli and Wu, Haoyuan and Wang, Xinyuan and Wang, Zekun and Yang, Junlin and Su, Hongjin and Chen, Jixuan and Chen, Junda and others},
  journal={arXiv preprint arXiv:2510.19488},
  year={2025}
}

@article{guan2026computer,
  title={Computer-Using World Model},
  author={Guan, Yiming and Yu, Rui and Zhang, John and Wang, Lu and Zhang, Chaoyun and Li, Liqun and Qiao, Bo and Qin, Si and Huang, He and Yang, Fangkai and others},
  journal={arXiv preprint arXiv:2602.17365},
  year={2026}
}

@article{koh2026generative,
  title={Generative Visual Code Mobile World Models},
  author={Koh, Woosung and Han, Sungjun and Lee, Segyu and Yun, Se-Young and Shin, Jamin},
  journal={arXiv preprint arXiv:2602.01576},
  year={2026}
}

@article{shen2026world,
  title={World-Model-Augmented Web Agents with Action Correction},
  author={Shen, Zhouzhou and Hu, Xueyu and Li, Xiyun and Fang, Tianqing and Li, Juncheng and Zhang, Shengyu},
  journal={arXiv preprint arXiv:2602.15384},
  year={2026}
}

@article{bai2025qwen3,
  title={Qwen3-vl technical report},
  author={Bai, Shuai and Cai, Yuxuan and Chen, Ruizhe and Chen, Keqin and Chen, Xionghui and Cheng, Zesen and Deng, Lianghao and Ding, Wei and Gao, Chang and Ge, Chunjiang and others},
  journal={arXiv preprint arXiv:2511.21631},
  year={2025}
}

@article{chen2025internalizing,
  title={Internalizing world models via self-play finetuning for agentic rl},
  author={Chen, Shiqi and Zhu, Tongyao and Wang, Zian and Zhang, Jinghan and Wang, Kangrui and Gao, Siyang and Xiao, Teng and Teh, Yee Whye and He, Junxian and Li, Manling},
  journal={arXiv preprint arXiv:2510.15047},
  year={2025}
}

@inproceedings{sun2025genesis,
  title={Os-genesis: Automating gui agent trajectory construction via reverse task synthesis},
  author={Sun, Qiushi and Cheng, Kanzhi and Ding, Zichen and Jin, Chuanyang and Wang, Yian and Xu, Fangzhi and Wu, Zhenyu and Jia, Chengyou and Chen, Liheng and Liu, Zhoumianze and others},
  booktitle={Proceedings of the 63rd Annual Meeting of the Association for Computational Linguistics (Volume 1: Long Papers)},
  pages={5555--5579},
  year={2025}
}

@article{zhao2025unified,
  title={Unified multimodal understanding and generation models: Advances, challenges, and opportunities},
  author={Zhao, Shanshan and Zhang, Xinjie and Guo, Jintao and Hu, Jiakui and Duan, Lunhao and Fu, Minghao and Chng, Yong Xien and Wang, Guo-Hua and Chen, Qing-Guo and Xu, Zhao and others},
  journal={arXiv preprint arXiv:2505.02567},
  year={2025}
}

@article{wang2025fostering,
  title={Fostering video reasoning via next-event prediction},
  author={Wang, Haonan and Liu, Hongfu and Liu, Xiangyan and Du, Chao and Kawaguchi, Kenji and Wang, Ye and Pang, Tianyu},
  journal={arXiv preprint arXiv:2505.22457},
  year={2025}
}

@article{wu2024atlas,
  title={Os-atlas: A foundation action model for generalist gui agents},
  author={Wu, Zhiyong and Wu, Zhenyu and Xu, Fangzhi and Wang, Yian and Sun, Qiushi and Jia, Chengyou and Cheng, Kanzhi and Ding, Zichen and Chen, Liheng and Liang, Paul Pu and others},
  journal={arXiv preprint arXiv:2410.23218},
  year={2024}
}

@inproceedings{zhang2025agentcpm,
  title={Agentcpm-gui: Building mobile-use agents with reinforcement fine-tuning},
  author={Zhang, Zhong and Lu, Yaxi and Fu, Yikun and Huo, Yupeng and Yang, Shenzhi and Wu, Yesai and Si, Han and Cong, Xin and Chen, Haotian and Lin, Yankai and others},
  booktitle={Proceedings of the 2025 Conference on Empirical Methods in Natural Language Processing: System Demonstrations},
  pages={155--180},
  year={2025}
}

@article{lu2025ui,
  title={Ui-s1: Advancing gui automation via semi-online reinforcement learning},
  author={Lu, Zhengxi and Ye, Jiabo and Tang, Fei and Shen, Yongliang and Xu, Haiyang and Zheng, Ziwei and Lu, Weiming and Yan, Ming and Huang, Fei and Xiao, Jun and others},
  journal={arXiv preprint arXiv:2509.11543},
  year={2025}
}

@article{myers1998hci-history,
  title={A brief history of human-computer interaction technology},
  author={Myers, Brad A},
  journal={interactions},
  volume={5},
  number={2},
  pages={44--54},
  year={1998},
  publisher={ACM New York, NY, USA}
}

@misc{claude-computer-use,
  author       = {{Anthropic}},
  title        = {Developing a Computer Use Model},
  year         = {2024},
  howpublished = {\url{https://www.anthropic.com/research/developing-computer-use}},
  note         = {Accessed April 27, 2026}
}

@article{gu2026generalization,
  title={Generalization in Online Reinforcement Learning for Mobile Agents},
  author={Gu, Li and Jiang, Zihuan and Chi, Zhixiang and Liu, Huan and Wang, Ziqiang and Yu, Yuanhao and Berseth, Glen and Wang, Yang},
  journal={arXiv preprint arXiv:2603.07432},
  year={2026}
}

@article{yao2026cgl,
  title={CGL: Advancing Continual GUI Learning via Reinforcement Fine-Tuning},
  author={Yao, Zhenquan and Huang, Zitong and Zeng, Yihan and Han, Jianhua and Xu, Hang and Feng, Chun-Mei and Ma, Jianwei and Zuo, Wangmeng},
  journal={arXiv preprint arXiv:2603.02951},
  year={2026}
}

@misc{cao2025androidlens,
  title = {AndroidLens: Long-latency Evaluation with Nested Sub-targets for Android GUI Agents},
  author = {Yue Cao and Yingyao Wang and Pi Bu and Jingxuan Xing and Wei Jiang and Zekun Zhu and Junpeng Ma and Sashuai Zhou and Tong Lu and Jun Song and Yu Cheng and Yuning Jiang and Bo Zheng},
  year = {2025},
  eprint = {2512.21302},
  archivePrefix = {arXiv},
  url = {https://arxiv.org/abs/2512.21302}
}

@misc{mu2025gui360,
  title = {GUI-360: A Comprehensive Dataset and Benchmark for Computer-Using Agents},
  author = {Jian Mu and Chaoyun Zhang and Chiming Ni and Lu Wang and Bo Qiao and Kartik Mathur and Qianhui Wu and Yuhang Xie and Xiaojun Ma and Mengyu Zhou and Si Qin and Liqun Li and Yu Kang and Minghua Ma and Qingwei Lin and Saravan Rajmohan and Dongmei Zhang},
  year = {2025},
  eprint = {2511.04307},
  archivePrefix = {arXiv},
  url = {https://arxiv.org/abs/2511.04307}
}

@misc{tang2026clawgui,
  title = {ClawGUI: A Unified Framework for Training, Evaluating, and Deploying GUI Agents},
  author = {Fei Tang and Zhiqiong Lu and Boxuan Zhang and Weiming Lu and Jun Xiao and Yueting Zhuang and Yongliang Shen},
  year = {2026},
  eprint = {2604.11784},
  archivePrefix = {arXiv},
  url = {https://arxiv.org/abs/2604.11784}
}

@misc{xu2026mobileagentv35,
  title = {Mobile-Agent-v3.5: Multi-platform Fundamental GUI Agents},
  author = {Haiyang Xu and Xi Zhang and Haowei Liu and Junyang Wang and Zhaozai Zhu and Shengjie Zhou and Xuhao Hu and Feiyu Gao and Junjie Cao and Zihua Wang and Zhiyuan Chen and Jitong Liao and Qi Zheng and Jiahui Zeng and Ze Xu and Shuai Bai and Junyang Lin and Jingren Zhou and Ming Yan},
  year = {2026},
  eprint = {2602.16855},
  archivePrefix = {arXiv},
  url = {https://arxiv.org/abs/2602.16855}
}

@misc{xue2026evocua,
  title = {EvoCUA: Evolving Computer Use Agents via Learning from Scalable Synthetic Experience},
  author = {Taofeng Xue and Chong Peng and Mianqiu Huang and Linsen Guo and Tiancheng Han and Haozhe Wang and Jianing Wang and Xiaocheng Zhang and Xin Yang and Dengchang Zhao and Jinrui Ding and Xiandi Ma and Yuchen Xie and Peng Pei and Xunliang Cai and Xipeng Qiu},
  year = {2026},
  eprint = {2601.15876},
  archivePrefix = {arXiv},
  url = {https://arxiv.org/abs/2601.15876}
}

@misc{openai2026codexagent,
  title        = {Codex: An Agent for Almost Everything},
  author       = {{OpenAI}},
  year         = {2026},
  month        = apr,
  day          = {16},
  howpublished = {\url{https://openai.com/index/codex-for-almost-everything/}},
  note         = {Product announcement. Describes Codex desktop app updates including computer use, built-in browser, image generation, memory, and plugins.}
}

@misc{yang2025ultracua,
  title = {UltraCUA: A Foundation Model for Computer Use Agents with Hybrid Action},
  author = {Yuhao Yang and Zhen Yang and Zi-Yi Dou and Anh Nguyen and Keen You and Omar Attia and Andrew Szot and Michael Feng and Ram Ramrakhya and Alexander Toshev and Chao Huang and Yinfei Yang and Zhe Gan},
  year = {2025},
  eprint = {2510.17790},
  archivePrefix = {arXiv},
  url = {https://arxiv.org/abs/2510.17790}
}

@article{falconui,
  title={Falcon-UI: Understanding GUI Before Following User Instructions},
  author={Shen, Huawen and Liu, Chang and Li, Gengluo and Wang, Xinlong and Zhou, Yu and Ma, Can and Ji, Xiangyang},
  journal={arXiv preprint arXiv:2412.09362},
  year={2024}
}

@article{qwenagentworld,
  title={Qwen-AgentWorld: Language World Models for General Agents},
  author={Zuo, Yuxin and Xiao, Zikai and Sheng, Li and Huang, Fei and Tu, Jianhong and others},
  journal={arXiv preprint arXiv:2606.24597},
  year={2026}
}

@article{xu2026mobileworldmodel,
  title={How Mobile World Model Guides GUI Agents?},
  author={Xu, Weikai and Huang, Kun and Feng, Yu and Li, Jiaxing and Chen, Yuhan and others},
  journal={arXiv preprint arXiv:2605.10347},
  year={2026}
}

@article{seerguard,
  title={SeerGuard: A Safety Framework for Mobile GUI Agents via World Model Prediction},
  author={Yu, Xue and Yuan, Bo and Yang, Pengjian and Zhao, Kailin and Hu, Hong and Feng, Junlan},
  journal={arXiv preprint arXiv:2607.15550},
  year={2026}
}

@article{patchworld,
  title={PatchWorld: Gradient-Free Optimization of Executable World Models},
  author={Bai, Jiaxin and Guo, Yue and Dong, Yifei and others},
  journal={arXiv preprint arXiv:2605.30880},
  year={2026}
}

@article{video2gui,
  title={Video2GUI: Synthesizing Large-Scale Interaction Trajectories for Generalized GUI Agent Pretraining},
  author={Xiong, Weimin and Gu, Shuhao and Ye, Bowen and Yue, Zihao and Li, Lei and Song, Feifan and Li, Sujian and Tian, Hao},
  journal={arXiv preprint arXiv:2605.14747},
  year={2026}
}

@article{phoneworld,
  title={PhoneWorld: Scaling Phone-Use Agent Environments},
  author={Tang, Zhengyang and Liu, Yuxuan and others},
  journal={arXiv preprint arXiv:2605.29486},
  year={2026}
}

@article{mobilegym,
  title={MobileGym: A Verifiable and Highly Parallel Simulation Platform for Mobile GUI Agent Research},
  author={Wu, Di and Hao, Ruitao and Wang, Haiyang and Wu, Shuzhe and Xiao, Han and others},
  journal={arXiv preprint arXiv:2605.26114},
  year={2026}
}

@article{weblica,
  title={Weblica: Scalable and Reproducible Training Environments for Visual Web Agents},
  author={Kar, O{\u{g}}uzhan Fatih and Bachmann, Roman and Gong, Yuting and Larsen, Anders Boesen Lindbo and Dehghan, Afshin},
  journal={arXiv preprint arXiv:2605.06761},
  year={2026}
}

@article{xiaomigui0,
  title={Xiaomi-GUI-0 Technical Report},
  author={Cao, Wanxia and Duan, Chengzhen and Fu, Pei and Gao, Pengzhi and others},
  journal={arXiv preprint arXiv:2606.31410},
  year={2026}
}

@article{mobileforge,
  title={MobileForge: Annotation-Free Adaptation for Mobile GUI Agents with Hierarchical Feedback-Guided Policy Optimization},
  author={Liu, Guangyi and Zhao, Pengxiang and Wu, Gao and Yin, Yiwen and Li, Mading and others},
  journal={arXiv preprint arXiv:2606.19930},
  year={2026}
}

\appendix
\section{Trajectory-to-Sample Conversion Examples}
\label{sec:sample_construction}

This appendix illustrates how an annotated trajectory is converted into training samples, using AgentNet as an example. An $N$-step trajectory $\tau = \left(g, s_1, a_1, s_2, a_2, \ldots\right)$ yields $N$ trajectory fine-tuning samples (one per step) and $N-1$ transition tuples $(s_t, a_t, s_{t+1})$, each of which instantiates one inverse dynamics sample and one forward dynamics sample. The abridged formats are shown below, where \texttt{<image>} denotes a screenshot input and the full action-format specification in the system prompt is omitted for brevity.

\textbf{Trajectory fine-tuning sample (step $S$).} The input contains the system prompt, the textual actions of all previous steps (past screenshots are excluded, as described in Section~\ref{subsec:experimental_setup}), the current screenshot $s_S$, and the task instruction; the target is the action of the current step.

{\small
\begin{verbatim}
[System]    You are a GUI agent. You are given a task and a screenshot
            of the screen. You need to perform a series of pyautogui
            actions to complete the task. <action-format specification>

[Assistant] # Step 1
            ## Action:
            Open the document "report.docx" in LibreOffice Writer.
            ...
[Assistant] # Step S-1
            ## Action:
            Click the "File" menu in the top-left corner.

[User]      <image: current screenshot s_S>
            # Task Instruction:
            Export the opened document as PDF and save it to Desktop.

            Please generate the next move according to the screenshot,
            task instruction and previous steps (if provided).

[Target]    # Step S
            ## Action:
            Click the "Export as PDF" button in the dialog.
            ## Code:
            ```python
            pyautogui.click(x=0.7318, y=0.4642)
            ```
\end{verbatim}
}

\textbf{Inverse dynamics sample.} The input contains the two consecutive screenshots; the target is the action that caused the transition.

{\small
\begin{verbatim}
[User]      You are a helpful assistant that infers actions from
            screen transitions.

            Previous screen:  <image: s_t>
            Current screen:   <image: s_{t+1}>

            Action format: <action-format specification>

            What action was performed?

[Target]    ## Action:
            Click the "Export as PDF" button in the dialog.
            ## Code:
            ```python
            pyautogui.click(x=0.7318, y=0.4642)
            ```
\end{verbatim}
}

\textbf{Forward dynamics sample.} The input contains the current screenshot and the action; the target is the next screenshot, supervised by the flow matching objective.

{\small
\begin{verbatim}
[User]      You are a helpful assistant that predicts the next screen
            state. Given the current screen, perform the move:
            ## Action:
            Click the "Export as PDF" button in the dialog.
            ## Code:
            ```python
            pyautogui.click(x=0.7318, y=0.4642)
            ```

            What is the next screen?
            <image: s_t>

[Target]    <image: s_{t+1}>
\end{verbatim}
}

\end{document}